\documentclass[a4paper,fleqn]{cas-sc}

\usepackage[authoryear]{natbib}

\usepackage{hyperref}       
\usepackage{url}            
\usepackage{booktabs}       
\usepackage{amsfonts}       
\usepackage{verbatim}      
\usepackage{nicefrac}       
\usepackage{microtype}      
\usepackage{xcolor}         
\definecolor{myorange}{RGB}{255,127,14}
\definecolor{mypink}{RGB}{175,64,123}
\usepackage{graphicx}
\usepackage{amsmath}
\newcommand\norm[1]{\left\lVert#1\right\rVert}

\def\tsc#1{\csdef{#1}{\textsc{\lowercase{#1}}\xspace}}
\tsc{WGM}
\tsc{QE}

\begin{document}
\let\WriteBookmarks\relax
\def\floatpagepagefraction{1}
\def\textpagefraction{.001}

\shorttitle{}    

\shortauthors{}  

\title [mode = title]{Self-supervised local learning rules learn the hidden hierarchical structure of high-dimensional data}  



%

\author[1]{Ariane Delrocq}
\fnmark[1]
\ead{ariane.delrocq@epfl.ch}
\author[1]{Wu S. Zihan}
\fnmark[1]
\ead{zihan.wu@epfl.ch}
\author[1,2]{Guillaume Bellec}
\author[1]{Wulfram Gerstner}
\cormark[1]

\affiliation[1]{organization={School of Life Science and School of Computer and Communications Sciences, EPFL},
            city={Lausanne},
            country={Switzerland}}
\affiliation[2]{organization={Machine Learning Research Unit, TU Wien},
            city={Vienna},
            country={Austria}}














\cortext[1]{Corresponding author}

\fntext[1]{Equal Contribution}


\begin{abstract}
The brain learns abstract representations of high-dimensional sensory input, but the plasticity rules that enable such learning are unknown. We study biologically plausible algorithms on the Random Hierarchy Model (RHM), an artificial dataset designed to investigate how deep neural networks learn the intrinsic hierarchical structure of high-dimensional data. We focus on two types of local learning rules that avoid both a long convergence time and the use of a symmetric error network. The first type uses direct feedback signals to approximate error propagation from the output layer. The second type uses layerwise self-supervised contrastive or non-contrastive loss functions that do not explicitly approximate errors at the output layer. 
  We show that all rules of the first type fail to solve the tasks of the RHM and trace this failure back to input-specific nonlinearities (`masking') that are implemented in full backpropagation and are essential for learning complex tasks. 
  However, algorithms of the second type are able to learn the hierarchical hidden structure of the  RHM tasks and are as data-efficient as supervised backpropagation training, while being compatible with known rules of synaptic plasticity in cortex.
\end{abstract}


\begin{highlights}
\item Local self-supervised learning rules learn hidden hierarchical data structure as data-efficiently as supervised learning with backpropagation.
\item Direct Feedback Alignment (DFA) is unable to efficiently learn hidden hierarchical data structure because the method ignores input-specific masking when approximating gradients.
\end{highlights}

\begin{keywords}
 Local learning rules \sep representation learning \sep hierarchical structure \sep self-supervised learning
\end{keywords}

\maketitle

\section{Introduction} \label{sec-intro}

Data samples collected in the real world often have a hierarchical structure. In vision, an `object' such as a bicycle generates very different pixel images depending on the viewing angle, the distance and inclination of the bicycle, as well as the illumination conditions \citep{Patel15}. The visual ventral stream of the brain extracts hierarchical representations of visual input, starting from simple Gabor-like filters in the early regions \citep{Hubel63} to face-sensitive and object-sensitive neurons in the Inferior Temporal cortex \citep{DiCarlo12}. Deep neural networks have been used to successfully learn the hierarchical representations of natural images and are able to predict recordings of neurons in the primate visual ventral stream \citep{Yamins16}. 

However, standard deep neural networks are optimized by backpropagation (BP), which is not biologically plausible \citep{Crick89, Lillicrap20}. First, to implement BP in the brain, an error network is needed to propagate the error backward using the exact same weights as the forward weights. This is known as the weight-transport problem. Second, a precisely timed signal needs to split BP into multiple phases: forward pass, error computation, backward pass, and weight updates. Numerous biologically plausible approaches have been proposed to approximate backpropagation \citep{Lillicrap16, Scellier17, Whittington17, Sacramento18, Meulemans21}. However, only a few of them resolve the two biological implausibilities of BP at the same time. Moreover, some of these  \citep{Scellier17, Whittington17, Meulemans21} also require the network to slowly converge to steady states in response to {\em each} input sample, before learning can take place. The long waiting time before learning is unlikely to be compatible with biology.  

To address bio-plausibility, methods inspired by Direct Feedback Alignment (DFA) \citep{Nokland16} use  long-range feedback connections to send error signals from the output layer to all hidden layers. Although such long-range connections have been successfully used in many studies \citep{Nokland16, frenkel19, Meulemans22, Srinivasan24}, these methods send flawed learning signals: the  matrix of  feedback weights is identical for all possible input samples, as opposed to BP where the effective feedback pathway depends on the derivative of the nonlinear transformation in the feedforward pathway.  In the case of ReLU networks, the nonlinearities translate into  partial, \emph{input-dependent} `masking' of connections in the backward pathway. 
Most empirical simulations that use variants of DFA have been conducted on simple tasks such as MNIST or CIFAR10, which do not need a deep network but are solvable by shallow networks with 1 or 2 hidden layers. Such tests on simple tasks bypass the potential problem of input-specific masking. The gap between DFA and BP further increases as the task complexity and network depth increase \citep{launay20}. As a first question, 
{\em we ask how methods with direct feedback signals} \citep{Nokland16, frenkel19, Meulemans22, Srinivasan24} {\em perform on difficult data sets where multiple processing layers are required.} 
To answer this question, we use the Random Hierarchy Model (RHM),  which generates a family of artificial datasets with an intrinsic hierarchical structure and adjustable degree of complexity \citep{Cagnetta24}. Using classical supervised learning with BP, it has previously been shown that the RHM task can be learned data-efficiently only by deep neural networks that have enough layers to capture the hidden hierarchies
\citep{Cagnetta24}. 

A second type of biologically plausible algorithm does not rely on ideas of DFA, but instead uses local, layer-specific loss functions for self-supervised learning \citep{Illing21, Halvagal23}. 
Standard self-supervised learning uses BP to learn meaningful representations in the output layer \citep{Oord19, Chen20, Caron21, Bardes22}. Biologically plausible local self-supervised objectives \citep{Illing21, Halvagal23} do not need BP, yet  show promising performance on datasets of intermediate levels of complexity. Due to the layerwise loss functions, the resulting weight updates follow local learning rules that satisfy all the biological plausibility criteria and do not require lengthy convergence of activities for each input sample.  
As a second question, {\em we ask how such local self-supervised learning rules perform on the RHM tasks.}
We find that  biologically plausible local self-supervised learning rules \citep{Illing21, Halvagal23} perform as well as global end-to-end self-supervised training. Moreover, the data efficiency of solving these tasks with local self-supervised learning approaches matches that reported for  classic supervised learning \citep{Cagnetta24}.

\section{Models}

\subsection{The Random Hierarchy Model (RHM)}

The RHM \citep{Cagnetta24} defines a
family of datasets with a controllable degree of complexity. In an RHM dataset, each sample is a codeword in the form of a sequence of $d = 2^L$ low-level features representing a hidden object identity (e.g., A, B, C ... in Figure 1A) where $L$ is the number of
encoding steps. Importantly, only the low-level features (level 1) are visible. The hidden high-level encoding steps generate a hierarchical data structure that needs to be `decoded' to correctly classify the data samples. The encoding levels are constructed recursively from top to bottom as follows.

\begin{figure}
    \centering
   \includegraphics[width=0.85\textwidth]{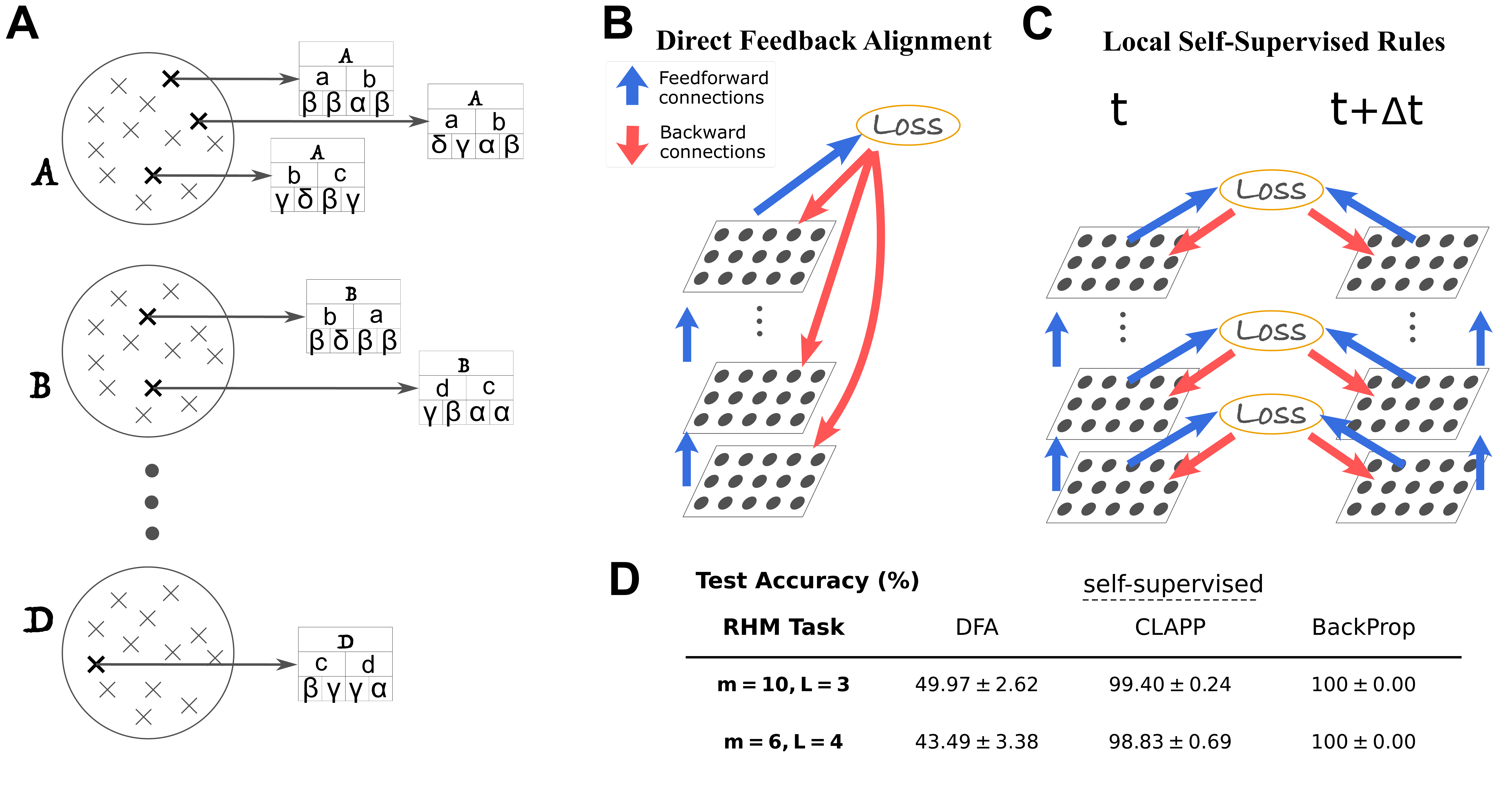}
    \caption{
	      {\bf The Random Hierarchy Model and local learning algorithms.}
          \textbf{A.} The RHM generates a family of hierarchical datasets. In a dataset, $n_c$ different top-level objects A,B,C, ...  are recursively encoded into strings composed of lower-level features. At each encoding step, each object (at the top level) or feature (at intermediate levels) is encoded by $m$ different synonyms; each synonym is a pair of 2 lower-level features, chosen from a  level-specific vocabulary of size $v$. Therefore a given object has many equivalent representations (all crosses for that object) linked by a shared hierarchical structure (shown for some example points). For example, with $m=3$, object A can be represented by "ab", "bc" or "dd" ; the feature "a" can be encoded by $\beta \beta$, $\delta \gamma$, or $\beta \alpha$, etc.  Note  that, after two steps of encoding, all objects are represented  by the same vocabulary of features $\alpha,\beta, \gamma, \delta$, but combined into different strings of length four. After $L$ encoding steps, the  many different strings of lowest-level features are the object encodings that serve as input to the neural `decoding' network. 
          \textbf{B.} Direct Feedback Alignment (DFA): the loss is computed at the output layer of a network with $L$ layers of neurons (black ellipses). To update the feedforward connections (blue arrows), error signals are directly propagated to each layer through a set of linear backward connections (red arrows).
          \textbf{C.} Local self-supervised rules: a self-supervised loss is computed at each layer of the network using the representations of paired (or batched) inputs. Paired inputs can be interpreted as a  model receiving input in two consecutive timesteps. To update the feedforward connections (blue arrows), error signals are computed from the corresponding layer-specific loss. There is no error signal backpropagation between layers.
          \textbf{D.} Test accuracy of different learning algorithms on two RHM tasks: $m=10, L=3$ and $m=6, L=4$, with $m=v=n_c$ (`maximal case'). Each task uses $P = 4 \cdot n_c v^L$ training data. The scaling factor for network width is set to $c_h = 5$. Values after $\pm$ are 95\% confidence interval computed from 8 runs with different randomly generated encoding rules and network initializations.
          Supervised BackProp serves as a control.
          \label{fig-intro} }
\end{figure}

The top level (numbered as level $L+1$)  has $n_c$ objects, written  as a string of length 1, e.g., "A", "B" or "C".  In the following, we refer to the symbols A,B,C as the features at level $L+1$. At each of the following encoding levels $l=L, \dots 1$, a level-specific vocabulary of $v$ new features is used. Each feature of level $l+1$ is represented by a pair of level $l$ features. Importantly, while each pair of level $l$ features can be uniquely decoded to reconstruct the feature at level $l+1$, a feature at level $l+1$ is mapped stochastically to one of its $m$ corresponding pairs of level $l$ features. Thus, in the example of \autoref{fig-intro} A, after the first level of encoding, object A is represented by any of the $m$ pairs \verb|ab|, \verb|bc|, \verb|dd|. Then, after the second level of encoding, the feature \verb|a| is encoded by, say, $\beta \beta$, $\delta \gamma$, or $\beta \alpha$, and \verb|b| by $\alpha \beta$, $\gamma \delta$, or $\beta \delta$. In this case, the strings $\beta\beta\alpha\beta$, $\gamma\delta\alpha\beta$, $\gamma\delta\beta\gamma$, and $\delta\gamma\gamma\delta$ are examples of encodings of object A. 


The dataset contains $P_{max} = n_c m^{d-1}$ such codewords. 
The choice of parameters $L$, $n_c$, $m$, and $v$ controls the structural complexity and size of the dataset.
As in \cite{Cagnetta24}, we study the maximal case, i.e., $m=v=n_c$. In this case, the full codeword cannot be predicted from any part of the codeword, thereby making self-supervised next-token prediction impossible. More generally, no small segment of the codeword can be used to predict the whole codeword or the original object -- analogously to the difficulty of uniquely detecting an object from a small set of its pixels. For a hierarchical depth $L$ the total size of the dataset is then $P_{max} = v^{2^L}$.
A training set of chosen size $P$ is randomly sampled from the full dataset. 

The Random Hierarchy Model was specifically designed to investigate backpropagation learning in deep networks \citep{Cagnetta24}, in a supervised framework. In the following, we explore whether biologically-plausible self-supervised algorithms or biologically-plausible approximations of BP are able to extract useful representations of objects encoded with an $L$-step RHM.

\subsection{Overview of biologically plausible learning rules}

Several families of synaptic plasticity rules have been proposed. Examples from three families will be compared in this study.

\paragraph{Variations of Hebb rules.} Learning in the brain is traditionally believed to be based on the Hebb rule \citep{Hebb49,Bi01,Bliss03,Malenka04}. Classic Hebb rules \citep{Hertz91,Gerstner02,Clopath10} describe the weight change as a combination of two factors, \emph{i.e.}, presynaptic activity and the state of the postsynaptic neuron. These rules can be used to perform Principal Component Analysis \citep{Oja82}, Independent Component Analysis \citep{Hyvarinen98},  k-means-clustering \citep{Grossberg76,Hertz91}, or Slow-Feature Analysis \citep{Sprekeler07}. However, none of these classical Hebbian synaptic plasticity has been shown to succeed in training deep neural networks to a level comparable to that of BP. As an example from this family of unsupervised learning rules, we consider a nonlinear Hebbian rule for ICA in this study \citep{Hyvarinen98,Hyvarinen00}.

\paragraph{Variations of equilibrium propagation.} 
Another family of local learning algorithms approximates BP through bidirectionally connected networks. During learning, the networks converge to equilibrium states while the output units are ``nudged'' or ``clamped'' to the target value \citep{Scellier17, Whittington17, Sacramento18, Meulemans21, Laborieux24, Salvatori24}. However, for each input sample, a long waiting time until convergence to equilibrium states is required for these algorithms. We do not consider this class of learning rules further, but note that the problem of waiting time is avoided in networks that send derivative signals \citep{Max24}.

\paragraph{Variations of feedback alignment.} A third family of algorithms approximate BP by simplifying the feedback pathway. Methods like Feedback Alignment (FA) \citep{Lillicrap16} addressed the weight transport problem by using random or trained feedback weights \citep{Akrout19, Max24}. Direct Feedback Alignment (DFA) \citep{Nokland16} further addressed the problem of network symmetry by completely replacing the feedback network by direct connections from the last layer to intermediate layers (\autoref{fig-intro} B). In this study, we use three examples  of this family of algorithms: DFA \citep{Nokland16}, Strong-DFC \citep{Meulemans22} and PEPITA \citep{Srinivasan24}.

\paragraph{Local self-supervised learning.} A fourth family of biologically plausible learning rules has been derived from loss functions for self-supervised learning applied separately at each layer \citep{Illing21,Halvagal23,Chen25}, avoiding the implausibility of BP through layers.
CLAPP \citep{Illing21} optimizes a contrastive objective for each layer (\autoref{fig-intro} C). 
The loss function of LPL (Latent Predictive Learning) \citep{Halvagal23} is the layerwise application of non-contrastive objectives, based on variance regularization.  
Variants of the forward-forward algorithms \citep{Hinton22forwardforward, Momeni25, Chen25} also optimize layerwise objective functions, resulting in local learning rules for the feedforward weights. Most of the above self-supervised rules can be considered  biologically plausible because they enable the following interpretation: 
a synapse in the cortex changes based on local factors such as the number and timing of spikes arriving at the synapse (first factor; input activity in the rate model case), the average voltage as well as voltage excursions caused by spikes (second factor; output activity or its derivative for rate models) \citep{Clopath10,Sjostrom01, Aceituno24}, as well as neuromodulators \citep{Pawlak10},   sometimes called  the "third" factor \citep{Gerstner18}.  In other words, the above rules fall into the framework of generalized three-factor rules. For contrastive self-supervised learning rules that compare two inputs, the third factor typically transmits a "sameness" signal that indicates the relationship between the two inputs \citep{Illing21}.
We now give details on the three self-supervised algorithms that we use in this study.

\begin{figure}
    \centering
   \includegraphics[width=0.9\textwidth]{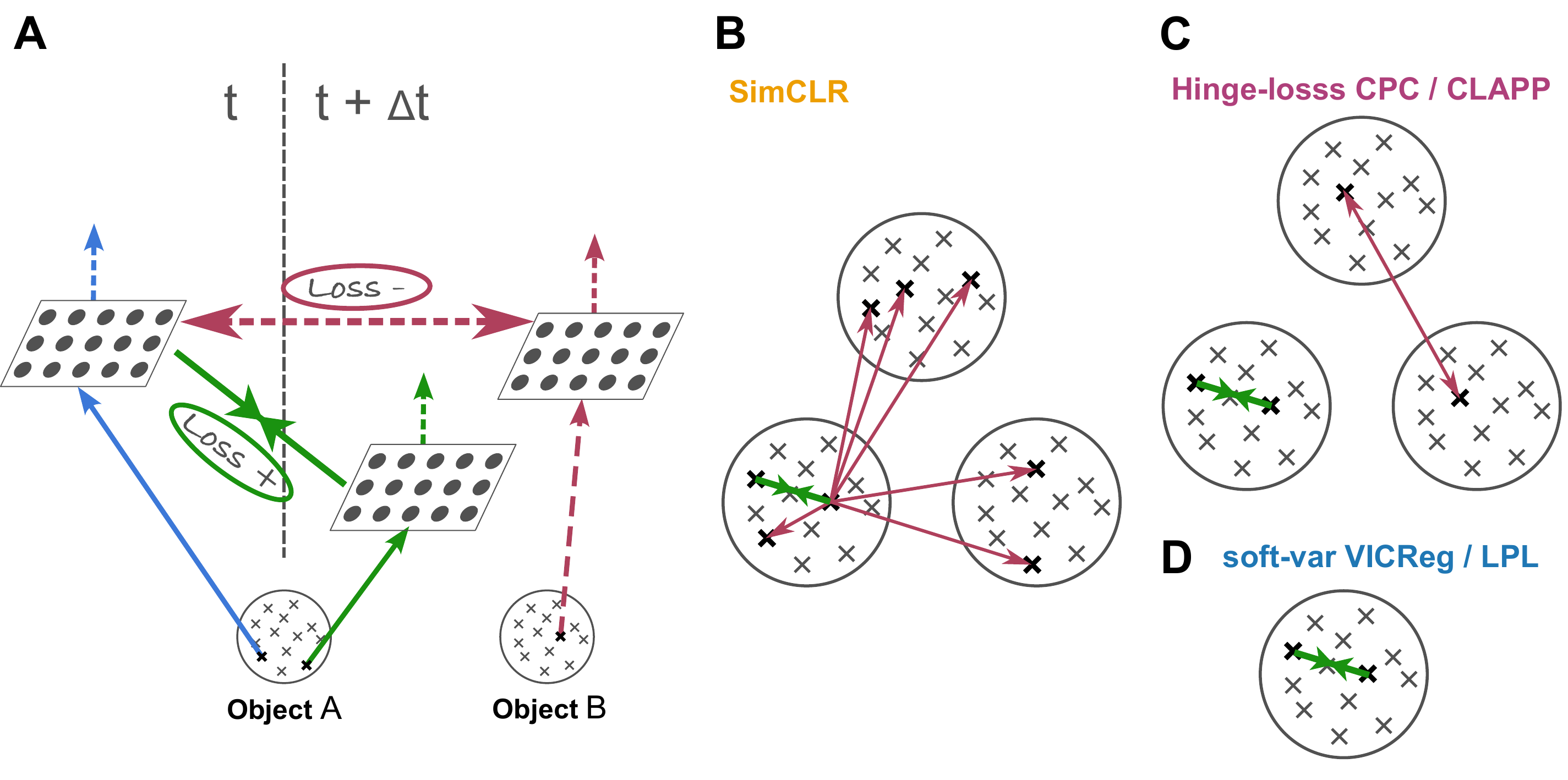}
    \caption{
	      {\bf Applying self-supervised algorithms to the Random Hierarchy Model.}
          \textbf{A.} Pairs of encodings of the same object (blue and green arrows) or different objects (blue and dashed magenta arrows) are used as input to the network at consecutive times $t$ and $t + \Delta t$. A local self-supervised loss is computed at each layer and then used to update weights of the same layer.
          \textbf{B, C and D.} Self-supervised algorithms compute the neuronal representations (activities) of an ANN for two representations of the same object and try to "pull together" these representations (green arrows).
          \textbf{B.} SimCLR maximizes the similarity of a "positive" pair and minimizes the similarities of all other possible pairs (with one input in common to the positive pair) \emph{via} a softmax function.
          \textbf{C.} Hinge-loss CPC / CLAPP chooses either a "positive" pair of samples (encodings of the same object) or a "negative" pair of samples (encodings of different objects) and updates the weights to move the two representations closer or further, respectively.
          \textbf{D.} LPL is a non-contrastive algorithm similar to VICReg \citep{Bardes22}: it does not have "negative" pairs but instead performs regularization in the neuron space. 
          \label{fig-ssl} }
\end{figure}

\subsection{Three self-supervised algorithms}
\label{sec-threeAlgos}
We work with three predictive self-supervised algorithms \citep{Illing21,Halvagal23,Chen20}. Here we review them and their application to the RHM framework. 
All three algorithms use a sameness signal as follows: to each data point $i$ in the batch, we assign another element $p(i)$ that encodes the same object as $i$. This assignment is called a `positive pair'. The algorithms compare the representations $\vec{z}_i$ and $\vec{z}_{p(i)}$ of the two data points to compute a predictive component in the latent space (\autoref{fig-ssl}). CLAPP and SimCLR also form negative pairs $(i, n(i))$ with data points $n(i)$ randomly sampled from the batch.

In real world data,  self-supervision signals (`sameness signal') arise if an agent fixates a moving 
object, or intentionally manipulates the object, and therefore obtains different views of the same object.
Since the agent is aware that the object identity remains unchanged, different views are naturally linked to each other as ``belonging to the same object''. The class identity, or name, of the object is never known, hence the name `self-supervision'. With data from the RHM, a "positive" pair is generated by picking any two encodings of the same object. Similarly, a "negative" pair is generated by picking two encodings of different objects (or any two encodings, see \autoref{clapp-nc}). The signal of ``not the same''  can be interpreted as self-awareness of performing a saccade from one object to the next.

Each of the three algorithms is applied in local BP-free fashion where feedforward weights are adapted by optimizing separately the loss function in each layer. The quality of the representations learnt by self-supervised models are tested by predicting the object identity with a linear decoder from the neuronal representations (output activations) \citep{Oord19, Chen20, Bardes22}.



\textbf{(i) SimCLR} \citep{Chen20} is a self-supervised algorithm designed to perform contrastive learning in the representation space by gradient descent on a softmax of the similarity $s_{i,p(i)}$ (scalar product) of the representations of pairs of inputs. The positive pair consists of two different encodings of the same object (in the original paper generated by two differently augmented versions of the same image), which is compared to the similarity of many other encodings. SimCLR uses transformations $\vec{z}_i$ and $\vec{z}_{p(i)}$ of the representations of two different input samples as a "positive" pair, \emph{i.e.} $s_{i,p(i)} = \vec{z}_i \cdot \vec{z}_{p(i)} / ||\vec{z}_i||~||\vec{z}_{p(i)}||$. 
Here, we use a simplified version of SimCLR (with details in \autoref{sec-suppl-ssl}). 

The SimCLR loss \citep{Chen20}, for a positive pair, is:
\begin{equation}
    \label{eq-simclr}
    \mathcal{L}^{\text{SimCLR}}_{i,p(i)} = -\log \frac{e^{s_{i,p(i)}}}{\sum_{k \neq i} e^{s_{i,k}}}
\end{equation}
The term in the numerator maximizes the similarity of ("pulls together")  the representation of two encodings of the same object and the term in the denominator "pushes apart" the representations of other encodings (where the sum runs over all encodings in the mini-batch); see \autoref{fig-ssl}B.
In practice, in a mini-batch we compute the summed losses of several positive pairs in a batch: $\mathcal{L}^{\text{SimCLR}} = \sum_{i} \mathcal{L}^{\text{SimCLR}}_{i,p(i)}$.
In classic SimCLR, the loss is applied only at the output layer of the neural network. Here, we apply the SimCLR loss separately to the representations in each layer of the network to obtain a layer-local BP-free version of SimCLR. 
Note that even without BP across layers, it is not straightforward to interpret the resulting algorithm as a bio-plausible learning rule: the use of the similarity of many pairs for one update would require recalling all the samples at the same time. SimCLR shares the same loss function as CPC \citep{Oord19}. The difference is that in the original CPC paper positive pairs were constructed from patches of the same image in an autoregressive model whereas in SimCLR positive pairs were constructed by pairs of augmented images.

 \textbf{(ii)  Hinge-loss CP/CLAPP} \citep{Illing21} is another variation of CPC,  adapted to avoid the biologically implausible simultaneous use of multiple "negative" pairs (as in SimCLR). Instead of a softmax, CLAPP uses the Hinge loss for a {\em single}  input pair  that is either positive ($j=p(i)$) or negative ($j=n(i)$) as indicated by the scalar $y_{ij}=\pm 1$:
 \begin{equation}
    \label{eq-CLAPP}
    \mathcal{L}^{\text{CLAPP}}_{i,j} = \max \left( 0, 1 - y_{ij}\vec{z}_i  W^{pred} \vec{z}_j \right )
\end{equation}
See  \autoref{fig-ssl}C; the dashed arrows in the figure indicate that the comparison is made for \emph{either} a positive or a negative pair (i.e., two samples encoding different objects).\footnote{In CLAPP, a \emph{negative} self-supervised signal is applied to randomly chosen pairs coming from  \emph{different} objects.  Appendix \ref{clapp-nc} shows that CLAPP performs similarly if negative pairs are chosen randomly without the condition that they must come from different objects.}
The similarity measure is the scalar product with learnable weights $W^{pred}$, as in CPC \citep{Oord19}.
In a mini-batch we make one (or a few) pair comparison per batch element: $\mathcal{L}^{\text{CLAPP}} = \sum_{i} \left(X_i\mathcal{L}^{\text{CLAPP}}_{i,p(i)} + (1-X_i) \mathcal{L}^{\text{CLAPP}}_{i,n(i)} \right)$ where $X$ is a binary mask assigning $i$ to either a positive or negative pair.
Hinge-loss CPC is mostly known in its layer-wise form, CLAPP \citep{Illing21}, which overcomes all the major implausibilities of CPC or SimCLR and can be interpreted as a generalized Hebbian learning rule acting locally at the synapse, with lateral predictions at the apical dendrites \citep{Aceituno24,Williams19,Urbanczik14}.

\textbf{(iii)  LPL} (Latent Predictive Learning) \citep{Halvagal23} is a non-contrastive self-supervised learning framework designed to respect the biological constraints of learning in the brain. 
LPL is originally layerwise but its loss can be seen as a variation of VICReg \citep{Bardes22} where the constraint on variance is softened; that is why we use the term `soft-var VICReg' to designate specifically the end-to-end version.
The predictive component of the loss is:
\begin{equation}
    \label{eq-LPL-pred}
    \mathcal{L}^{\text{LPL, pred}}_{i,p(i)} =  \frac{1}{2}\left( \vec{z}_i - \vec{z}_{p(i)}\right)^2
\end{equation}
Similar to the other algorithms, positive pairs are chosen among the encodings of the same object, but there are no negative pairs: this algorithm does not "push apart" the representations of unrelated input. To avoid representational collapse, LPL uses regularizations in the neuron space rather than the sample space.
The two regularization terms needed to compensate for the absence of negative pairs are computed over the batch, for the neurons $f,g$:

\begin{equation}
    \label{eq-LPL-var}
    \mathcal{L}^{\text{LPL, var}} = \left\langle -\log(\sigma_f^2) \right\rangle_f
\end{equation}
\begin{equation}
    \label{eq-LPL-decor}
    \mathcal{L}^{\text{LPL,decorr}} =  \left\langle \text{Cov}(f,g)^2\right\rangle_{f\neq g}
\end{equation}

where $\sigma_f^2$ is the variance of neuron $f$ computed over a batch of input, and $\text{Cov}(f,g)$ is the sample covariance of neurons $f$ and $g$, computed over the batch.
The total loss gathers the three components with weighting coefficients $c_1$ and $c_2$:
\begin{equation}
    \label{eq-LPL}
    \mathcal{L}^{\text{LPL}} =  \sum_{i} \mathcal{L}^{\text{LPL,pred}}_{i,p(i)} + c_1 \mathcal{L}^{\text{LPL, var}} + c_2 \mathcal{L}^{\text{LPL,decorr}}
\end{equation}

\section{Results}

We train deep neural networks on data arising from the family of datasets generated by the RHM. 
The string of lowest-level features is used as an input to the neural network. Following \citep{Cagnetta24}, our neural networks are convolutional neural networks (CNNs) with as many hidden layers as there are hierarchical levels in the RHM dataset and an architecture (kernel size and stride) matching the dataset (both equal to 2). The size (number of features) of the hidden layers is set to $c_h v^2$, where $v^2$ is the dimension of the encoding of one higher-level feature and $c_h$ is a scaling factor for network width. Due to the convolutional architecture, the dimension of the neural activity thus shrinks at each layer.

We first evaluate three published biologically-plausible learning rules: DFA \citep{Nokland16}, CLAPP \cite{Illing21}, and LPL \citep{Halvagal23}. We train CNNs on two RHM tasks with levels of complexity $L = 3$ and $L=4$ with small training sets, and we evaluate the linear classification accuracy for each on a test set of previously unseen data points. Supervised training using DFA \citep{Nokland16} fails to solve the task, while the two local self-supervised rules, CLAPP and LPL, solve the tasks at a performance level that nearly matches that of supervised training with BP (\autoref{fig-intro} D). The technical details of training can be found in Appendix A.



In the first two subsections, we investigate an important limitation of DFA \citep{Nokland16} and variations thereof \citep{Meulemans22,Srinivasan24} that hinders these learning rules from solving the RHM tasks \citep{Cagnetta24}. We mathematically define the concept of input-specific masking and then demonstrate, through simulations, its importance for approximations of the BP algorithm. 
In the third and fourth subsection, we further study the learning capacity and learnt representations of local self-supervised learning rules on RHM tasks. In the final subsection, we study the learning capacity of more naive, unsupervised biologically-plausible learning on RHM tasks.

\subsection{Input-specific masking in BP approximation}

We study an $L$-layer neural network with feedforward weight matrices $\{W_l\}_{l=1}^L$. Given input $x_\mu$, the activity at each layer $l$ is given by the vector $z_{\mu, l} = \rho(h_{\mu, l}) = \rho(W_l z_{\mu, l-1})$, with $z_{\mu, 0} = x_\mu$. We use vector notation and apply the neuronal nonlinearity $\rho(.)$ component-wise. We denote the output layer error $e_\mu = \partial \mathcal{L}/\partial z_{\mu, L}$. The update of BP for each batch input $B$ can be written as \citep{Goodfellow16}:
\begin{equation}
    \label{eq:bp}
    \frac{\partial \mathcal{L}}{\partial W_l} = \sum_{\mu \in B} [\rho'(h_{\mu,l}) \odot W_{l+1}^T  \rho'(h_{\mu, l+1})\odot  \cdots \odot W_L^T \rho'(h_{\mu, L}) \odot e_\mu ] z_{\mu, l-1}^T
\end{equation}

where $\odot$ is elementwise multiplication, and the computation in the bracket is performed from right to left in a non-commutative manner. Here, we define the derivatives of the forward activations $\{\rho'(h_{\mu,l})\}$ as the {\bf \em input-specific masking}. They are input-specific because the values depend on input $x_\mu$. We used the term `masking' because the derivatives are often 0 or very close to 0 for popular choices of activation function $\rho$, such as ReLU, Sigmoid, or Tanh. In particular, for ReLU networks, the derivatives can only be 0 and 1, thereby effectively becoming masks. 

Feedback Alignment (FA) \citep{Lillicrap16} replaces the transposed matrices $W_l^T$ with fixed matrices $B_l$, but still applies input-specific masking $\rho'(h_{\mu, l})$ for each layer:
\begin{equation}
    \label{eq:per_sample}
    \frac{\partial \mathcal{L}}{\partial W_l} = \sum_{\mu \in B} [\rho'(h_{\mu, l}) \odot B_{l+1}  \rho'(h_{\mu, l+1}) \odot \cdots \odot  B_L \rho'(h_{\mu, L})\odot e_\mu] z_{\mu, l-1}^T 
\end{equation}

Direct Feedback Alignment (DFA) \citep{Nokland16} removes all the intermediate masking terms $\rho'(h_{\mu, l})$ and projects the  error vector $e_\mu$ measured at the output back to layer $l$. Using the same notation as in FA,  we get
\begin{equation}
    \label{eq:no_mask}
    \frac{\partial \mathcal{L}}{\partial W_l} = \sum_{\mu \in B} [\rho'(h_{\mu, l}) \odot B_{l+1} \cdots  B_L e_\mu] z_{\mu, l-1}^T 
\end{equation}

In the original DFA \citep{Nokland16}, a single fixed matrix is used to project $e_\mu$. Here we write it as a product $B_{l+1} ... B_L$ to better emphasize the suppression of masking terms compared to Eqs. \ref{eq:bp} and \ref{eq:per_sample}. Since these feedback weights are fixed in DFA, writing the feedback matrix as a product of matrices does not affect the learning rule.

\subsection{Lack of input-specific masking limits learning with DFA}

 In FA and DFA, learnable forward weight matrices $W_l$ first try to align with preset feedback weights $B_l^T$ and then optimize the loss function in the neighborhood of the parameter space \citep{Refinetti21}. Additional (pre-)training of feedback weights to align with forward weights further improves performance \citep{Meulemans22, Srinivasan24, Max24, Akrout19}.
 More generally, for alignment to become possible, the feedback weights $B_l$ must be well chosen at initialization of learning.

  \begin{figure}
    
    \centering
   \includegraphics[width=\linewidth]{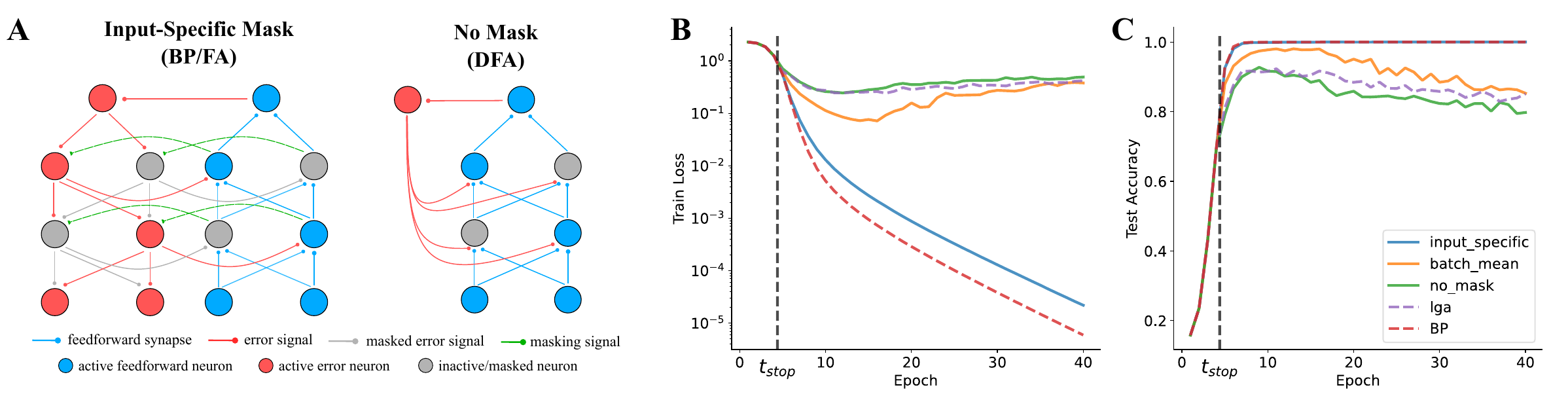}
   
    \caption{\label{fig-mask}
	      {\bf Input-specific masking is critical in BP-approximations}. \textbf{(A)} Illustration of different maskings in a ReLU network. In BP or FA, the inactive feedforward neurons (which have $\rho' = 0$) mask backward error flow in the corresponding feedback network. The error signal from these neurons becomes 0 (`masked error signal'). \textbf{(B)} Epoch-averaged training loss and \textbf{(C) }accuracy on the test dataset as the network is trained by different learning rules. RHM data was generated with $v=m=n_c=10$ and $L = 3$. A total of $4\cdot n_c v^l$ data are used for training and 10000 for testing. A deep convolutional network with 3 hidden layers of 100 neurons ($c_h$ = 1) is trained with cross-entropy loss and BP until the training accuracy (with sliding average window of 20 batches) hits 80\%, which is indicated by the vertical black dashed line at $t_{\rm STOP}$. 
          At this point BP is stopped and backward weights are fixed as the transpose of feedforward weight at $t_{\rm STOP}$. 
          Training is  then continued using approximate BP with three different maskings: input\_specific (equation \ref{eq:per_sample}), no\_mask (equation \ref{eq:no_mask}), or batch\_mean (equation \ref{eq-batch_mean}). The line 'lga' represents the simulation in which the optimized linear gradient approximation $B_l^{\rm lga}$ (equation \ref{eq-lga} \& \ref{eq-update-lga}) is used as the direct feedback weight.
    }
\end{figure}

Therefore,
to eliminate the specific choice of feedback weights $B_l$ as potential causes of failure,
we choose feedback weights not randomly as suggested in the original paper \citep{Nokland16}, but as the transpose of `almost ideal' feedforward weights. To achieve a near-optimal choice of feedback weights, we first train the network with BP until the network reaches a training accuracy of 80$\% $ which defines the total pre-training time $t_{\rm STOP}$. Then we use the resulting feedforward weights to set the feedback weights $B_l =W_l^T(t_{\rm STOP})$ and continue training
weight matrices $W_l$ with or without input-specific masking while keeping $B_l$ fixed. When there is no masking, we effectively perform DFA as in equation \ref{eq:no_mask}. With input-specific masking, we effectively perform FA as in equation \ref{eq:per_sample}. 
In order to further study the importance of input-specificity, we also studied two further learning rules called `batch-averaged masking' and `linear gradient feedback' (\autoref{sec-suppl-mask}), two approximations of input-specific masking, as middle-ground points between FA and DFA.

As masking effects manifest only in networks with at least two hidden layers, we train a 3-layer network to solve several RHM tasks, each with $L=3$ hierarchical levels. We found that input-specific masking enables networks to solve RHM tasks, while removal of masking,  batch-averaged masking, or linear gradient feedback all prevent successful learning (Figure \ref{fig-mask}).  Similar results are observed with a 4-layer network trained on the RHM task with 4 encodling levels  (\autoref{sec-suppl-mask}, Figure \ref{fig-mask-l4}).

We know from earlier work with supervised learning that narrow networks find a compressed representation, whereas wide networks often find solution types based on random features \citep{Jacot18, chizat19, Lee19}. To illustrate the importance of masking, we therefore used in Figure \ref{fig-mask} a narrow network that has just the minimal number of hidden neurons per layer ($c_h = 1$) needed to efficiently solve the task.
In a wider network, it could be possible for the network to mostly use the last one or two hidden layers to solve the task, whereas the earlier layers do not learn meaningful features. In this case, the effect of masking would be alleviated. This is confirmed in simulation of a 4-layer network with $c_h=5$ trained on RHM task with four levels of hierarchy (\autoref{sec-suppl-mask}, figure \ref{fig-mask-l4}). Nevertheless, even in a wide network, the learning rules without input-specific masking do not reach the same level of performance as the ones with proper masking.
 Overall, this shows that DFA does not work on the family of RHM datasets, even with near-optimal (rather than random) feedback weights.

This result generalizes to recent biologically plausible learning rules that use long-range connections from the last layer: Strong-DFC \citep{Meulemans22} and PEPITA \citep{Srinivasan24}. When naively adapted to the RHM task with 3 levels of hierarchy, both fail  to learn the task (\autoref{sec-suppl-mask}, Figure \ref{fig-pepita-dfc}).

\subsection{Local self-supervised algorithms learn the hidden intrinsic structure of data from the RHM }

We use the three local self-supervised algorithms reviewed in Section \ref{sec-threeAlgos} – two contrastive and one non-contrastive – to perform representation learning with a CNN of $L$ layers. After training, we evaluate the learned representations by using a linear classifier to predict the object identity from the representation of the last layer. 
All three local self-supervised algorithms solve the task.

For RHM tasks \citep{Cagnetta24}, the interesting question is not only the learnability for local self-supervised algorithms, but also the amount of data required to solve the task. 
If a model is powerful enough to capture the structure of the data, a training set size that is a polynomial of the critical number  $D^*=n_c v^L$  is sufficient to learn the task \citep{Cagnetta24}. We call such a model `data-efficient'. In this case, the representations exhibit the same hierarchical structure as the data which allows the model to generalize easily to unseen samples from a minority of training samples. However, it is also possible to successfully solve the task by building representations that are only separable in a very high-dimensional space and that therefore need a large amount of data to be formed. The critical size $D^*=n_c v^L$ corresponds to the dataset size at which (i) correlations between the object identity and low-level features become theoretically apparent, and (ii) an invariance of the representations to exchanges of synonyms is found during supervised learning \citep{Cagnetta24}. In the following, we study the criterion $D^*$ also for local self-supervised learning rules, and show that (ii) applies as well in this case.

\begin{figure}
    \centering
   \includegraphics[width=\linewidth]{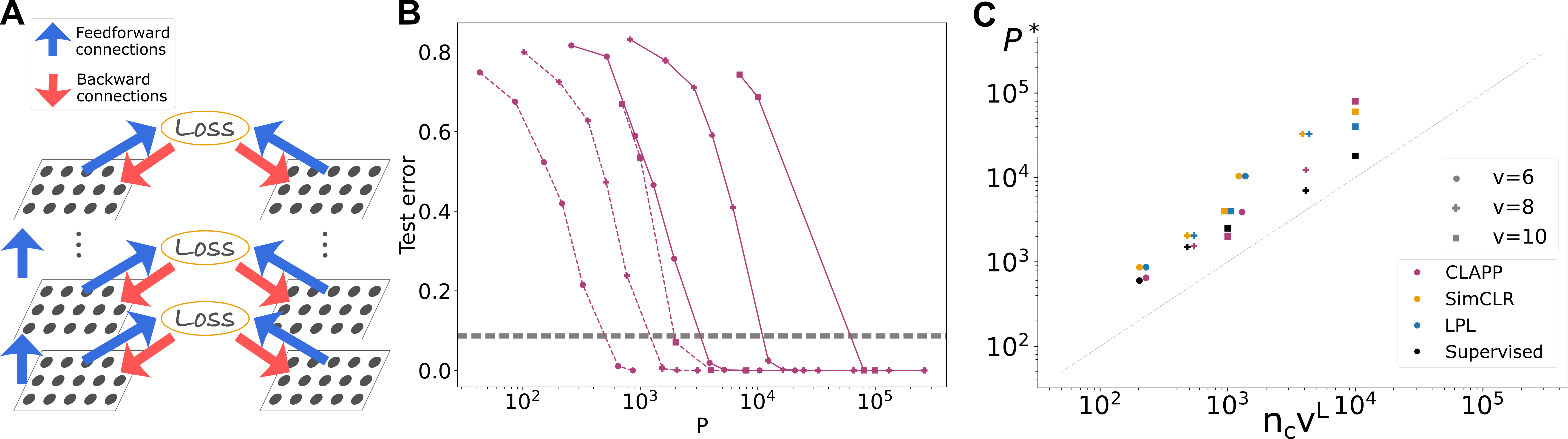}
    \caption{
	      {\bf  Local self-supervised algorithms solve RHM task data-efficiently}
          {\bf A}. Schematic of forward and backward flow with the local self-supervised algorithm. 
          {\bf B}. Test error of CLAPP learning rule as a function of training data size $P$ in different RHM configurations: $n_c = m = v=6, 8, 10$ and $L =2, 3$ (represented by different marker shapes and linestyles). 
          {\bf C}.
          The minimum dataset size $P^*$ required to learn and generalize to the whole dataset (linear decoding error $\leq 10\%$ of random error; threshold shown as grey dashed line in {\bf B}), as a function of $D^*=n_c v^L$, for the RHMs with parameters $L=2, 3$ and $n_c = m = v=6, 8, 10$ (shape of symbols; color presents type of loss). Corresponding $L$-layer CNNs with $c_h=5$ are trained with either CLAPP (pink), LPL (blue) or layerwise SimCLR (yellow) successfully form representations of the dataset with high data efficiency. $P^*$ scales with the dimension of the hierarchical structure $D^*$  rather than with the dataset size  $P_{max} = v^{2^L}$. In black, the supervised results from \citep{Cagnetta24} are shown as reference. Overlapping points are slightly shifted laterally to show all data.
          \label{fig-lw} }
\end{figure}

Specifically, we train CNNs on RHM datasets of varying complexity ($L=2, 3$ and $v=6, 8, 10$), each with different dataset sizes $2D^*$, $4D^*$, $8D^*$, $10D^*$, $20D^*$ for all three self-supervised algorithms. Among these values, we report the minimum amount of data needed to achieve a test error $\leq 10\%$ of random error. Here, for a more consistent comparison, we use the same CNNs with $c_h = 5$ as in supervised training from \cite{Cagnetta24}.

Our local self-supervised algorithms are able to build representations of the data that are useful --in the sense that linear readout is possible (\autoref{fig-lw} B). We find that all three algorithms can solve the tasks with a number of data points that scales with $D^*$, indicating that they successfully capture the hierarchical structure of the RHM data (\autoref{fig-lw} C). These self-supervised algorithms are slightly less data-efficient than the supervised training reported by \citet{Cagnetta24}, but the scaling law looks comparable. We also notice that there is no difference in performance between the two contrastive algorithms and the non-contrastive one, consistent with the claim of their equivalence \citep{Garrido23}.

The scaling with our local bio-plausible self-supervised rule is the
same as the one found if self-supervised loss functions are applied only at the output, with end-to-end training using BP (Appendix \ref{sec-ete}). Importantly, this means that for the type of hierarchical data generated by the RHM, the coordination of the learning of different layers by BP is neither necessary nor useful. This is highly relevant to real-world tasks, which are believed to be intrinsically hierarchical \citep{Patel15,Mhaskar17,Mossel18,Cagnetta23}.

\subsection{Local self-supervised rules extract hierarchical structures}

\begin{figure}
\hspace{0.01\textwidth}
  \centering
  \vspace{0pt}
  \includegraphics[width=\linewidth]{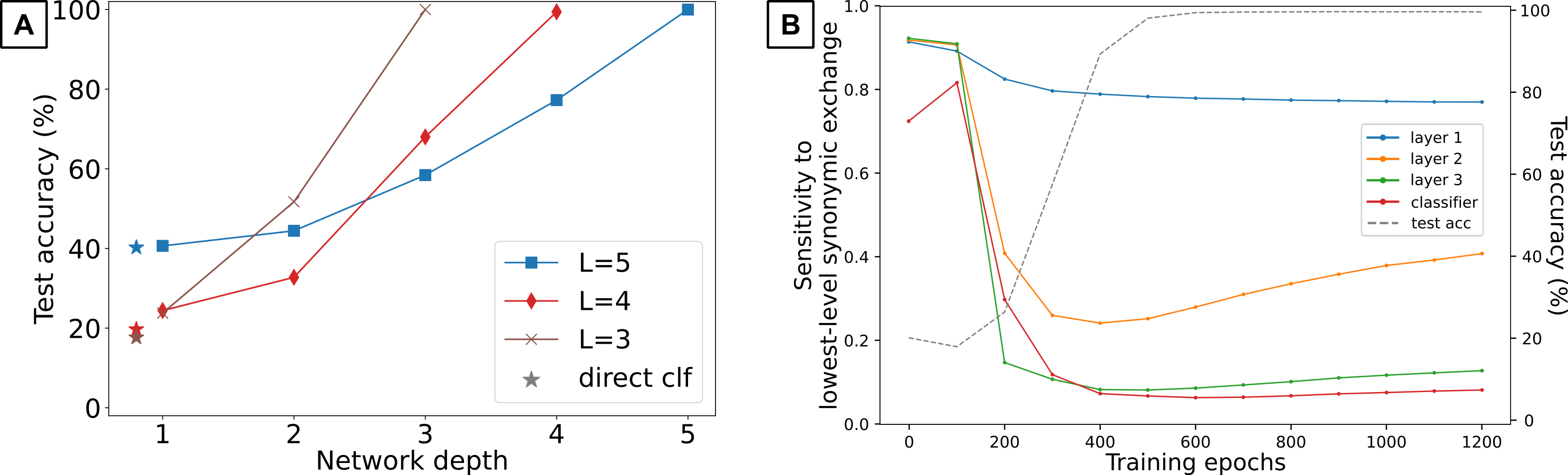}
  \caption{{\bf  CLAPP extracts the hierarchical structure of the RHM.}
       {\bf A}.  Data from an RHM of depth $L=5$ (with $m=v=3$ equivalent codes per feature, blue) or $L=4$, ($m=6$, red)  or  $L=3$ ($m=8$, brown) is decoded by CLAPP in networks of 5,  4 or 3 layers, respectively. Test accuracies measured by linear classification on the output of networks of different depths.  Asterisks: linear classification applied to the raw input of length $2^L$.
       {\bf B}. Sensitivity (vertical axis) of representations in layer 1, 2, 3 (resp. blue, orange, green) and of linear readout  (red) to switches between synonyms at the input layer. 
       The data was generated by an RHM model of depth $L=3$. As a function of training time (horizontal axis), the representation in the neural network becomes 
       invariant (low sensitivity) for different synonyms of the first level
       in layers 2, 3, and output, but not in layer 1. This switch parallels the increase of accuracy of the linear readout (grey dashed).
          \label{fig-clapp}}
\end{figure}

In order to understand how self-supervised layer-local algorithms learn deep hierarchical data of up to 5 layers, we select CLAPP, the strongest (for both visual and RHM tasks) of the bio-plausible algorithms tested above. Previously, CLAPP has been shown \citep{Illing21} to work on deep networks for representation learning in various modalities (images, videos, sound). For RHM datasets of depth $L=3,4,5$ we train CNNs of depth $\leq L$ (still with constant number of features per layer equal to a few $v^2$) with the local, bio-plausible CLAPP learning rule and find  (\autoref{fig-clapp}A) that, for an RHM of depth $L$, CLAPP leads to near-perfect representations in layer $L$ of the network when trained on a small portion of the dataset (with $P^* \approx \text{a few} D^*$).
However, networks that are shallower than the dataset structure fail at solving the RHM task in a data-efficient way.
In other words, networks of depth matching that of the hierarchy, but not shallower, can capture the dataset's hierarchical structure, in agreement  with the results found with supervised BP training  \citep{Cagnetta24}.

As a result of the match between the RHM hierarchy and the trained deep network, the representation of synonyms at different levels (of the RHM hierarchy) is stable in different layers (of the trained neural network). For example, in an RHM of depth 3, we consider that the same high-level object is represented by two different synonyms at the input level.
If, for a pair of synonyms, the RHM synonym-generation process differs only in the final codeword generation step (level $l=$1), then the representation in layer one of the trained network is highly sensitive to the exchange of these synonyms, whereas the representation  in layers 2, 3 or in the read-out layer is much less sensitive  (\autoref{fig-clapp}B).
Analogous statements are valid if the synonym-generation process differs at an intermediate encoding step (Supplementary information D, and Fig. \ref{fig-ss}).
This shows that
depth is used to decompose the input string presented to the input layer of the neural network in a step-wise fashion uncovering its  hierarchical composition. 
Moreover, \autoref{fig-clapp}B shows that the bulk of the insensitivity of the representations to the exchange of synonyms is acquired simultaneously in all (higher) layers.

\subsection{Naive  (non-predictive) Hebbian learning rules}
To understand whether the self-supervised signal is necessary to extract the hierarchical structure of RHM data, we also probed a non-predictive Hebbian algorithm. 
It is known that Hebbian learning rule can implement Independent Component Analysis (ICA) \citep{Hyvarinen98,Hyvarinen00}. Hebbian implementations of ICA  \citep{Illing19} have shown that ICA can solve simple tasks with a shallow network of one hidden layer \citep{Illing19}. Here we apply a Hebbian ICA  algorithm and use it to extract features from the RHM data. 

\begin{figure}
\centering
  \centering
  \vspace{0pt}
  \includegraphics[width=0.5\linewidth]{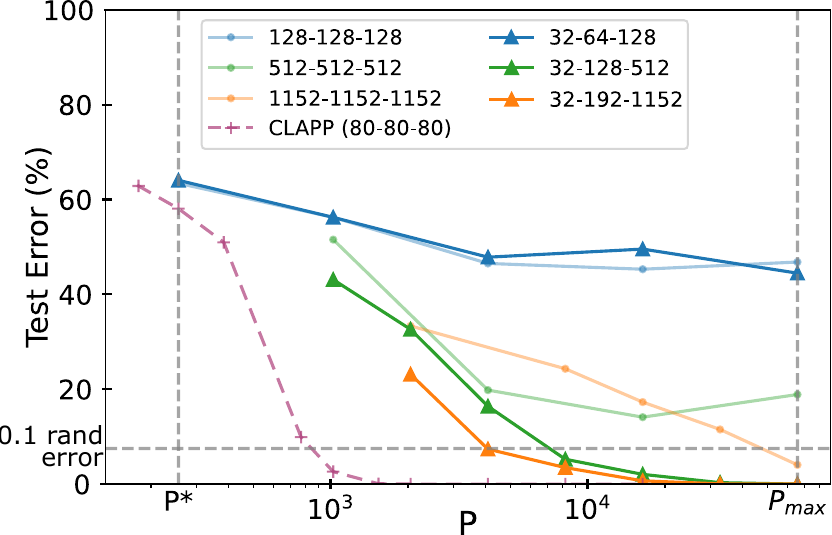}
  \caption{{\bf ICA needs both many data samples and many neurons to solve the RHM task.}
          Test error as a function of the number of samples in the training set, for different CNN sizes (with constant number of features 
          (pale colors and dot markers), or with increasing number of features per layer, dark colors and triangular markers). As a reference, CLAPP performance is shown in pink. 
          The RHM task for $L=3$ and $m=v=n_c=4$ is only solved by ICA networks with wide output layers (green/orange) and when training with a big dataset size $P$.
          \label{fig-ica}}

\end{figure}

We use the same one-dimensional CNN architecture as in the self-supervised algorithms: 
a convolutional kernel is trained to perform an ICA on an input segment of length 2. Therefore input dimensionality is $2v$, and the maximum number of output features for each ICA is $2v$. In order to match the hidden dimensionality to that of our networks used in self-supervised methods, we follow \citep{Illing19} and apply a bootstrapping strategy for the first layer: for each kernel we train multiple ICAs, each with an output dimension $2v$, and concatenate their outputs as the kernel output.
In our multi-layer network, the output of each layer is also passed through a ReLU activation. Feedforward weights in each layer are trained by ICA, on $P$ segments of length $s$ randomly sampled from the $P$ input words. 
The ICA representations in the final layer are tested by training a linear classifier to identify objects from the neural activities.
We use an RHM with $L=3$ and $m=v=n_c=4$; test errors are shown in \autoref{fig-ica}.

ICA was unsuccessful in building useful representations of the RHM for CNNs with sizes similar to those used for our end-to-end and local SSL algorithms ($5v^2=80$ features per layer).
However,  systematically increasing the number of neurons from one hidden layer to the next and again to the output layer made the task learnable by ICA.
Wide networks where all hidden layers had the same width performed considerably worse.
Since ICA requires large amounts of training data,  the test set for the linear decoder may overlap with the training data used for representation learning, which explains the values for $P=P_{max}$. This can at most make the test error look better than it would be on never-seen data, which only strengthens our point that ICA is insufficient to learn the structure of the RHM.

The fact that the unsupervised learning algorithm needs both high-dimensional representations and huge amounts of data 
to solve the RHM tasks confirms that self-supervision (in our case the "sameness" signal, necessary to detect correlations of representations with object identity) is crucial to learning high-dimensional hierarchical data: ICA cannot disentangle the data points without a enormous increase of dimensionality, which requires both (i) the formation of high-dimensional representations and (ii) exposure to most of the dataset.

\section{Discussion}

Our results reveal a fundamental limitation of methods that approximate backpropagation through long-range feedback, such as DFA: the lack of input-specific masking prevents them from learning hierarchical representations  of RHM tasks, even when feedback weights are well-initialized. This limitation becomes critical in deeper networks where multiple layers of non-linear transformations are needed.

In contrast, we have shown that SimCLR, LPL and CLAPP, both in a layer-wise version and with end-to-end BP training, are capable of extracting the intrinsic structure of the RHM, while ICA is not. It is likely that this result applies to a broader class of self-supervised algorithms, as long as they (i) leverage the correlations of input features across  pairs of encoding strings and 
(ii) are applied to  neural networks that are sufficiently deep to extract the hierarchical representations.

Although depth is necessary to build useful representations of the RHM data, local learning rules are sufficient to leverage the available depth and build hierarchical representations.
Notably, both CLAPP and LPL were designed to be compatible with physiological constraints on learning mechanisms in the brain and are nonetheless powerful enough to be data-efficient (in the sense that they learn hierarchical data from a relatively small fraction of the data).
Importantly,
the "sameness" signal that is explicitly or implicitly used in all the self-supervised algorithms that we investigated seems to be crucial for 
hierarchical, data-efficient learning, since ICA, a Hebbian rule without a predictive component, is data-hungry -- or fails altogether. Algorithms such as SFA \citep{Sprekeler07} are conceptually midway between ICA and LPL, but have not been tested on this task.
In biological terms, in a sensory processing pathway such as vision, our findings imply that a simple self-supervised learning signal, with a local weight update rule, is sufficient for each brain area to build useful hierarchical representations of the input, without the need for detailed feedback from higher areas.

At this stage, our conclusions remain at an abstract, conceptual level. First, our simulation experiments are limited to rate models and `simple' plasticity rules on CNNs, which disregards the complexity of cell morphology, spiking behavior (timing, bursting...), lateral and backward connectivity, short-term plasticity, noise in neuronal pathways etc. 
Second, all simulation experiments were performed on artificial data which enabled us to analyze the scaling behavior of algorithms with data efficiency. Yet, it is unclear how realistic the assumptions of the RHM are with respect to data arising in domains of vision, music, or language that form the primary motivation for the RHM \citep{Cagnetta24}. 
Future work would benefit from additional \emph{families} of datasets incorporating domain-specific knowledge while enablinf exploration of scaling laws with the fraction of data  necessary for training. 

The success in the synthetic task of algorithms  that combine broad plausible concepts of predictive learning, plasticity-inducing predictive apical dendritic input (CLAPP), and scalar "sameness" signal opens the possibility that evolution may have discovered and implemented functionally
similar mechanisms for real-world learning. 
Moreover, related insights and local learning rules are potentially also relevant for non-standard computing hardware \citep{Momeni25,Goltz21,Maryada25}.

\appendix

\section{Supplementary information: Technical details}

\subsection{The RHM}

\paragraph{Correlations in the RHM: the maximal case \emph{vs} incomplete datasets}
As discussed in the original paper, the encoding of the RHM can generate two different types of datasets, either full code ("the maximal case") or incomplete code. \citep{Cagnetta24}

All our work is based on \emph{full} datasets,  characterized by $n_c = m = v$. In this maximal case, at any level of the hierarchy, all $v^d$ possible combinations of features are part of the dataset (they encode an object).
There are $P_{max} = v^{(2^L)}$ different codewords or samples in such a dataset.
During the hierarchical dataset construction, "the random choice induces correlations between low- and high-level features." \citep{Cagnetta24} The value $P=D^*$ corresponds to the dataset size at which such correlations between the encodings' features and object identity become statistically apparent. Theoretical analysis of this statement can be found in \citep{Cagnetta24}. In addition, they show that this is also the dataset size at which deep neural networks with supervised training start learning representations that are invariant to the exchange of "synonyms" at different levels.

In the \emph{incomplete} code, 
not all of the $v^d$ potential codewords are part of the dataset. For example,  $\alpha\beta\beta\gamma$ would be unused. Note that this is different from sampling $P$ codewords for the training set, in which case the other codewords of the dataset still encode an object. In the incomplete code, some combinations of lower-level features are \emph{not} a codeword. This is the case assumed in token prediction \citep{Cagnetta2024b},  which relies on the correlations between the different features of an encoding.
"In the intermediary case $1 < m < v^{s-1}$, the distribution of the inputs has itself a hierarchical structure" \citep{Cagnetta24}: such correlations (for example, where $\alpha \alpha$ might be followed more often by $\alpha \beta$ than by $\beta \alpha$) only arise in \emph{incomplete} datasets.

It is important to note that it is the encoding-to-object correlations that are at play in our work. That is why unsupervised algorithms cannot use the correlations to learn the structure of the dataset. 
Token prediction is strictly impossible for full code.

\subsection{Training}
All networks are trained by stochastic gradient descent on the losses described in section \ref{losses} with the Adam optimizer and mini-batch size 128 for CLAPP/Hinge-loss CPC, 512 for soft-var VICReg, and values in $\{32, 128, 512\}$ for other algorithms (not optimized due to computational costs). The above applies to both representation learning and linear decoder training. The learning rate is set to $2\cdot 10^{-4}$ for CLAPP/Hinge-loss CPC,  $5\cdot 10^{-4}$  for LPL/soft-var VICReg, and $3\cdot 10^{-4}$ for SimCLR (incl. layerwise). These values were chosen after trying various learning rates in $\{5\cdot 10^{-3}, 1\cdot 10^{-3}, 5\cdot 10^{-4}, 3\cdot 10^{-4}, 2\cdot 10^{-4}, 1\cdot 10^{-4}, 5\cdot 10^{-5}\}$ on a few RHM datasets for each algorithm.

The training of each CNN was performed on 1 NVIDIA V100 GPU with 32GB memory.
Representation learning (training of the CNN) lasted up to a week for the bigger datasets (less than 15 minutes for the smaller datasets), while training the decoder lasted no more than a few hours (except for very big datasets and when the early stopping criterion was never reached, \emph{i.e.} classification failed).
Because of the high computational cost of training with big datasets, networks were trained for at most 20'000 epochs, and were sometimes interrupted earlier at arbitrary epochs if evaluation at the current stage was successful already. A few longer experiments suggest that continuing training (much) longer might allow some networks for slightly smaller values of $P^*$ than are reported in our results to achieve successful representations as well. However, even after up to 50'000 epochs in some cases, $P^*$ remains in the same order of magnitude.

The success of representation learning is tested by linear decoding of object classes from the representations of the encodings. 
The training set for the linear decoder is the same set with \emph{the same} $P$ samples as the training set for representation learning; the test set contains $\min(100'000, P_{max} - P)$ samples not present in the training set.
As in \citep{Cagnetta24}, the decoder is trained until the training loss consistently falls below $0.001$ (or for a maximum of 5'000 epochs, if that condition is never reached; in these cases the resulting test accuracy was insufficient). Learning rate is set to $2\cdot 10^{-4}$.

\paragraph{ICA}
ICA does not take the form of standard ANN training.
The ICA algorithm is implemented directly using the FastICA algorithm from the scikit-learn package with default hyperparameters.
To match the structure of a 1D convolutional neural network, we follow the procedure in \citep{Illing19}. More specifically, we split the input to each layer into segments of the length 2, same as the kernel size. The segments of all inputs are then used for FastICA to extract independent components. A linear rectifier (ReLU) is then applied on the output of FastICA at each layer.

Each hidden layer has an input dimension of $2 \cdot D_{l-1}$, where $D_{l-1}$ is the dimension of the previous layer. As a result, in the case of $D_l > 2 \cdot D_{l-1}$, directly applying FastICA would not be possible. Under such situation, we bootstrap the input and train multiple FastICA, each with an output of $2 \cdot D^{l-1}$ dimension. For each FastICA, we sampled $p$ input segments, where $p$ is the size of the training dataset. Then, we concatenate the linearly rectified output of these FastICA to achieve the desired dimension $D_l$.

The layers are trained in a greedy way (the second layer extracts independent components from the already fitted components of the first, etc). Fitting was performed on the CPU of Apple M2 Pro chip.

\paragraph{Deep networks \label{deep}}
In our \autoref{fig-clapp}, the CNN of $L$ layers is trained until the top-layer loss falls consistently below $0.001$. 
Then, to probe whether earlier layers solve the task already, we evaluate independently by linear classification the neuronal representations in each layer of the CNN.
Since there is no feedback across layers in the CLAPP model, lower layers learn independently of higher layers, and this is equivalent to training and evaluating a network of each depth $l \leq L$.

Note that at the output of the last ($L$-th) layer of the CNN, the "spatial" dimension of the convolution is reduced to one; the representation consists of the vector of features (neuronal activities) of this layer. For \autoref{fig-clapp}, the representations at lower layers are obtained by concatenating these feature vectors for the different locations of the spatial map (flattening the representations).

\subsection{Self-supervised algorithms\label{losses}}
\label{sec-suppl-ssl}
The self-supervision signal that makes the strength of the self-supervised algorithms is the "sameness" signal, indicating that two encodings represent the same object and therefore allowing to make predictive  pairs. 
In terms of biological interpretation, this can be envisioned as  harnessing the continuity/permanence of objects and concepts. CLAPP also uses the opposite, negative signal; this is discussed in Appendix \ref{clapp-nc}.

In the next subsections, we describe in details our implementation of the different self-supervised algorithms. For this purpose, we consider a batch of representations $\mathbf{z} \in \mathrm{R}^{(B\times D)}$ where $B$ is the batch size and $D$ the dimension of the representation. 
Likewise, we define the restriction $\mathbf{z}^{o} \in \mathrm{R}^{(n_o\times D)}$ of the vector $\mathbf{z}$ containing only the $n_o$ elements of the batch that are encodings of object $o$.

\subsubsection{SimCLR}
In the original paper \citep{Chen20}, the representations $\mathbf{z}$ are transformed by a non-linear MLP before used to compute the self-supervised objective function. Here, we omit this non-linear transformation and directly use the representation to compute the self-supervised loss function. We found this simplified version is already sufficient to learn the RHM tasks.

\begin{equation}
    \label{eq-simclr-detail}
    \mathcal{L}^{\text{SimCLR}} = \frac{1}{\sum_o n_o (n_o - 1)}
    \sum_{o}
    \sum_{i=1}^{n_o}
    \sum_{j=1 (j \neq i)}^{n_o}
    - \log 
    \frac
    {e^{sim\left( \mathbf{z}^o_{i},  \mathbf{z}^o_{j}\right)}}
    {\sum_{k=1}^B  e^{sim\left( \mathbf{z}^o_{i},  \mathbf{z}_{k}\right)}}
\end{equation}
where $sim\left( \mathbf{z}^o_{i},  \mathbf{z}_{k}\right)$ is set to $-10^8$ if the $k$-th element of $\mathbf{z}$ is the $i$-th element of $\mathbf{z}^o$.
For our experiments $sim(x,y)$ is simply the scalar product of vectors $x$ and $y$.

\subsubsection{LPL}
The three components of the LPL loss are defined as follows.
\begin{equation}
    \mathcal{L}^{\text{LPL, pred}} =  \frac{1}{B}\sum_{i=1}^B \left( \mathbf{z}^o_i - \mathbf{c}^o_i\right)^2
\end{equation}
We denote $\mathbf{c}^o$ the vector of "positive" samples for object $o$. This is simply a vector $\mathbf{c^o} \in \mathrm{R}^{(n_o\times D)}$ that contains the same representations as $\mathbf{z}^o$ in randomly shuffled order.
Recall that there are no negative samples.

For the expression of the variance and decorrelation components, we use superscript to denote batch index and subscript to denote feature index. As before, $M$ is the number of features in the neuronal representation. 
\begin{equation}
    \mathcal{L}^{\text{LPL, var}} =  \frac{1}{M} \sum_{f=1}^{M} - \log \left( \frac{1}{B-1} \sum_{i=1}^B \left( \mathbf{z}_f^i - \mathbf{\bar{z}}_f \right)^2 + \epsilon \right)
\end{equation} 
Here $\epsilon = 10^{-4}$ is a small constant for numerical stability.
For the variance part of the loss above as for the decorrelation part below,  $\bar{z}$ is the vector average of features computed over the batch, with \emph{stop gradient} applied,
treating $\bar{z}$ as a constant, outside of the gradient computation graph (\emph{i.e.} the gradient of the loss does not compute with respect to $\bar{z}$).
\begin{equation}
    \mathcal{L}^{\text{LPL,decorr}} =  \frac{1}{M(M-1)}\sum_{f=1}^{M} \sum_{g \neq f} \left(\frac{1}{B-1}\sum_{i =1}^B  \left( \mathbf{z}_f^i - \mathbf{\bar{z}}_f \right) \left( \mathbf{z}_g^i - \mathbf{\bar{z}}_g \right) \right)^2
\end{equation} 
The three components are combined as follows
\begin{equation}
    \label{eq-LPL-detail}
    \mathcal{L}^{\text{LPL}} = \mathcal{L}^{\text{LPL, pred}} + c_1 \mathcal{L}^{\text{LPL, var}} + c_2 \mathcal{L}^{\text{LPL,decorr}}
\end{equation}
where we use $c_1 = 1$ and $c_2 = 10$.

\subsubsection{CLAPP}

 \begin{align}
    \label{eq-CLAPP-detail}
    \mathcal{L}^{\text{CLAPP}} = \sum_o
    \sum_{k=1}^5
    \biggl(
    &\frac{1}{\sum \alpha_i} \sum_{i=k+1}^{n_o}
    \alpha_i \max \left( 0, 1 - \mathbf{z}^o_i  W^{pred} \mathbf{z}^o_{i-k} \right ) \\
    &+ \frac{1}{\sum (1-\alpha_i)} \sum_{i=k+1}^{n_o} (1 - \alpha_i)  \max \left( 0, 1 + \mathbf{z}_i  W^{pred} \mathbf{c}^o_{i-k} \right )
    \biggr)
\end{align}
Here, opposite to the LPL case, we denote $\mathbf{c}^o$ the vector of "negative" samples.
In the main paper, we take $\mathbf{c} \in \mathrm{R}^{((B-n_o)\times D)}$ to be the vector of all batch elements not in $\mathbf{z}^o$. In Appendix \ref{clapp-nc}, to avoid the selection of encodings of \emph{different} objects only, we instead take $\mathbf{c^o} \in \mathrm{R}^{(B\times D)}$ that contains the same representations as $\mathbf{z}$ in randomly shuffled order (and $\mathbf{c^o}$ is re-defined with a random permutation for each object $o$).

Here the use of $y \in \{ \pm 1 \}$ is for clarity replaced by the gating $\alpha\in \{ 0, 1 \}$ which is randomly chosen to be $0$ or $1$ with equal probabilities for each comparison $i$.
The $\alpha_i$'s and $(1-\alpha_i)$'s are summed over the same indices as $i$ runs over, but the summation indices are not written for better legibility.
Also, both $\mathbf{z}^o$ and $\mathbf{c}$ are truncated (along the batch dimension) to the length of the smaller of the two of them (usually $\mathbf{z}^o$), and $n_o$ takes the value of this length (usually unchanged).
$W^{pred}$ is trained by a simple Hebbian rule as in \citep{Illing21}.

Note that in this implementation of the CLAPP loss, a few artifacts are inherited from the expression of the standard CLAPP loss for image patches. They are theoretically irrelevant and removing them would not alter the results.  Notably, $\sim5$ comparisons are made per element in the batch, each time omitting $k$ elements of the batch. This does not change the "sample efficiency" argument as this argument does not involve the number of epochs/comparisons, but the number of different samples in the training set. The shifting by $k$ of the samples in $\mathbf{z}$ replaces the use of a shuffled version of $\mathbf{z}$. Also, the average positive and negative losses are summed (instead of taking the overall average), which may introduce a small random imbalance, but it statistically cancels out over batches.

\section{Supplementary information: input-specific masking}
\label{sec-suppl-mask}

\paragraph{Batch-averaged masking.} In this learning rule, the mask is computed as the mean of $\rho'(h_{\mu, l})$ for each batch, $\langle \rho'(h_{\mu, l+1})\rangle_\mu$:
\begin{equation}
    \label{eq-batch_mean}
    \frac{\partial \mathcal{L}}{\partial W_l} = \sum_{\mu \in B} [\rho'(h_{\mu, l}) \odot B_{l+1}  \langle \rho'(h_{\mu, l+1})\rangle_\mu \, \odot \cdots \odot B_L \langle\rho'(h_{\mu, L})\rangle_\mu \odot e_\mu] z_{\mu, l-1}^T 
\end{equation}

\begin{figure}
    
    \centering
   \includegraphics[width=13cm]{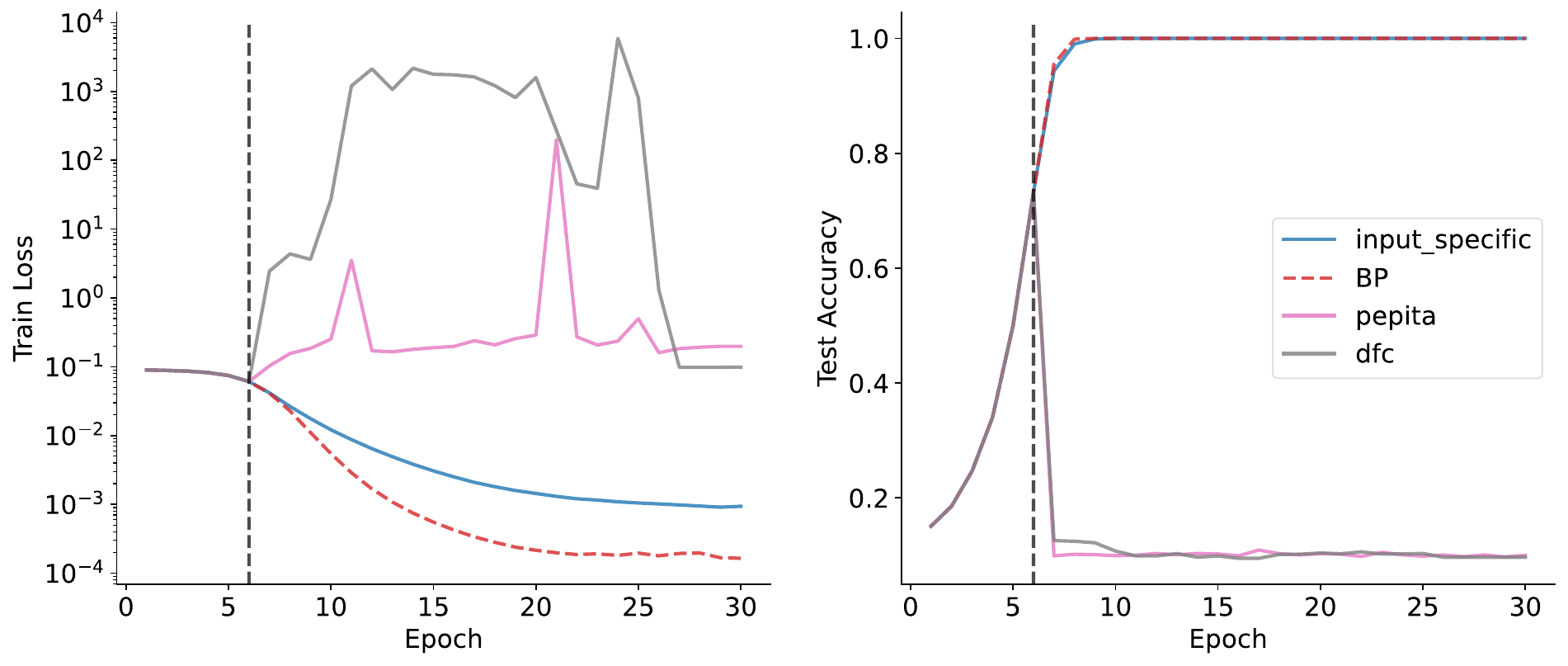}
   
    \caption{\label{fig-pepita-dfc}
	      {\bf Other biologically plausible learning with long-range feedback also fail}. The setup is the same as figure \ref{fig-mask}, except that we use mean squared error as loss and remove bias in the network.
    }
\end{figure}

\paragraph{Linear gradient approximation.} The linear gradient approximation is defined such that we determine the exact full-batch gradient direction of BP $\sum_\mu \frac{\partial \mathcal{L}}{\partial h_{\mu, l}}$ at layer $l$
and then choose backward weights such that they minimize difference 
to the full-batch gradient
\begin{equation}
\label{eq-lga}
B_l^{\rm lga} = \operatorname{argmin}_{\tilde{B}_l}
\sum_\mu \norm{\tilde{B}_l e_\mu - \frac{\partial \mathcal{L}}{\partial h_{\mu, l}} }_2^2 \, .
\end{equation}
where the superscript `lga' stands for Linear gradient approximation at time $t_{\rm STOP}$.
After optimizing $B_l^{\rm lga}$ across all batches (i.e., full-batch), we update the feedforward weights using:
\begin{equation}
\label{eq-update-lga}
\frac{\partial \mathcal{L}}{\partial W_l} = \sum_{\mu \in B} [\rho'(h_{\mu, l}) \odot B_l^{\rm lga} e_\mu] z_{\mu, l-1}^T 
\end{equation}
Choosing feedback weights $B_l^{lga, T}$ is similar to  'no-mask' training, but can be seen as  an alternative choice of  'ideal' initialization. The Linear gradient approximation is different from the batch-averaged masking (equation \ref{eq-batch_mean}) because the minimization takes the error $e_\mu$ into account:
\begin{align*}
    B_l^{\rm lga} &= \operatorname{argmin}_{\tilde{B}_l}
\sum_\mu \norm{\tilde{B}_l e_\mu- W_{l+1}^T  \rho'(h_{\mu, l+1}) \odot \cdots \odot W_L^T \rho'(h_{\mu, L}) \odot e_\mu }_2^2 \\
&= \operatorname{argmin}_{\tilde{B}_l}
\sum_\mu  \norm{\left[\tilde{B}_l - W_{l+1}^T  \rho'(h_{\mu, l+1}) \odot \cdots \odot W_L^T \rho'(h_{\mu, L}) \odot \right] e_\mu}_2^2
\end{align*}
We note that the norm $\mu$ (i.e. all patterns in the full batch) includes the local output error $e_\mu$. In contrast, batch-averaged masking does not depend on $e_\mu$:

\paragraph{Variations of DFA: PEPITA and DFC}
In figure \ref{fig-pepita-dfc}, we tested two biologically plausible learning rules with long-range feedback: Strong-DFC \citep{Meulemans22} and PEPITA \citep{Srinivasan24}. We evaluated them using the same task and network parameters as in Fig. \ref{fig-mask}. However, we use mean squared error as loss and remove bias in the network, because the PEPITA algorithm is only written for bias-free network trained by MSE loss. For PEPITA, we initialize the direct top-down feedback weights the same as for DFA: the product of the transpose of `almost ideal' feedforward weights. For Strong-DFC, because they have a learning rule for feedback weights, we initialize the feedback weights randomly and train the feedback weights as in \citep{Meulemans22}. Note that neither of the two algorithms claim to approximate BP. We tried to stay as close as possible to the original set of hyperparameters. 

\begin{figure}
    
    \centering
   \includegraphics[width=13cm]{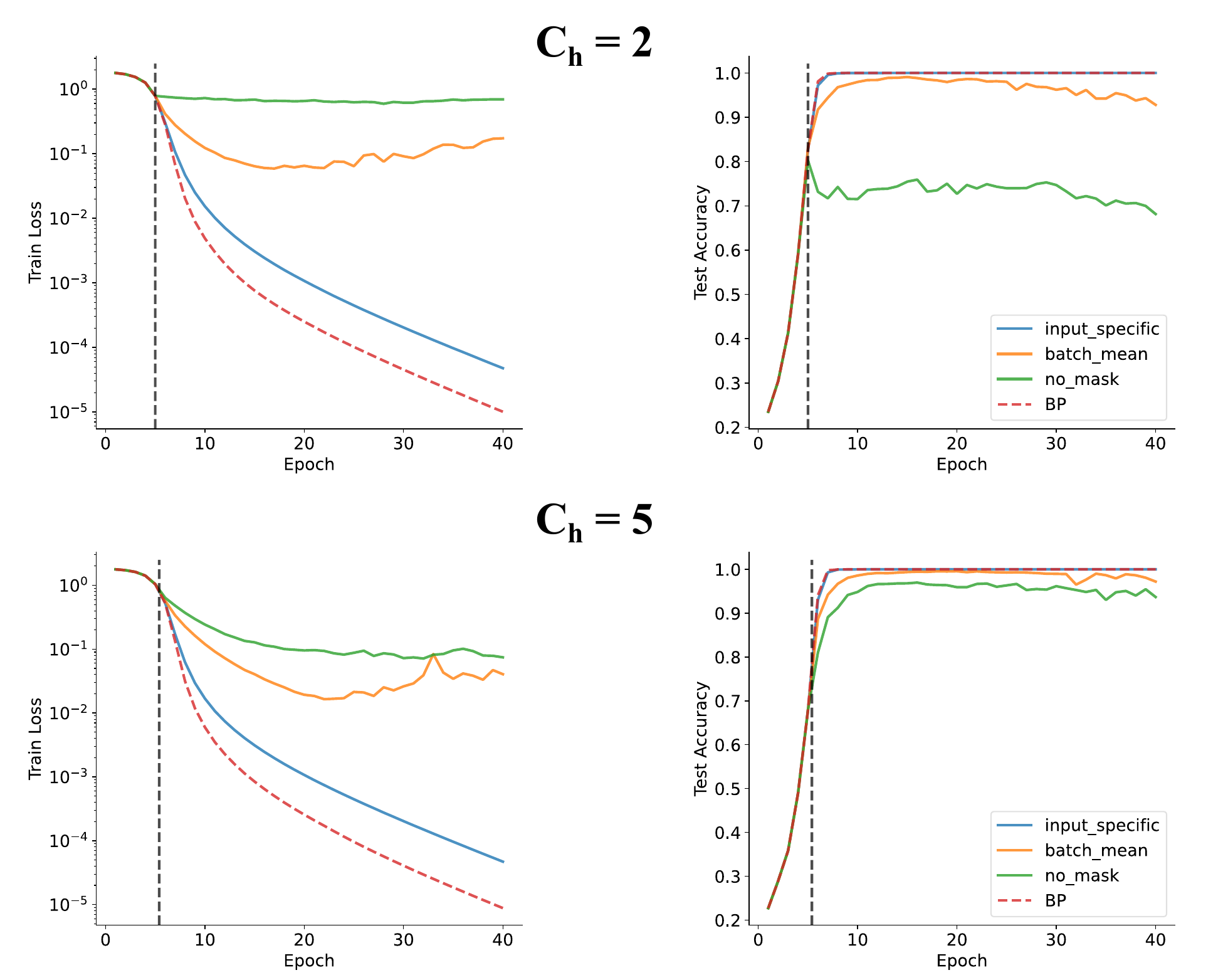}
   
    \caption{\label{fig-mask-l4}
	      {\bf RHM with four levels of hierarchy, otherwise as in Figure \ref{fig-mask}}. Epoch-averaged training loss (left) and accuracy on the test dataset (right) as the network is trained by different learning rules. RHM data was generated with $v=m=n_c=6$ and $L = 4$. A total of $4\cdot n_c v^l$ data are used for training and 10000 for testing. Deep convolutional network with 4 hidden layers are trained with cross-entropy loss and BP until the training accuracy (with sliding average window of 20 batches) hits 80\%, which is indicated by the vertical black dashed line. Thereafter feedback weights were kept fixed. \textbf{Top}: narrow network with 72 neurons ($c_h = 2$) per layer; \textbf{Bottom:}  wide network with 180 neurons ($c_h = 5$) per layer.
    }
\end{figure}

\paragraph{Narrow versus wide networks}
In figure \ref{fig-mask-l4}, we demonstrate the importance of input-specific masking for 4 layer convolutional networks trained on the RHM task with 4 levels of hierarchy. With a narrow network with $c_h = 2$, we observe the same qualitative results as with the  narrow network in the main text trained for 3 levels of hierarchy. On a wider network ($c_h=5$), we find less impact from input-specific masking.

\section{Supplementary information: end-to-end self-supervised algorithms solve RHM tasks}
\label{sec-ete}

\begin{figure}
    \centering
   \includegraphics[width=11cm]{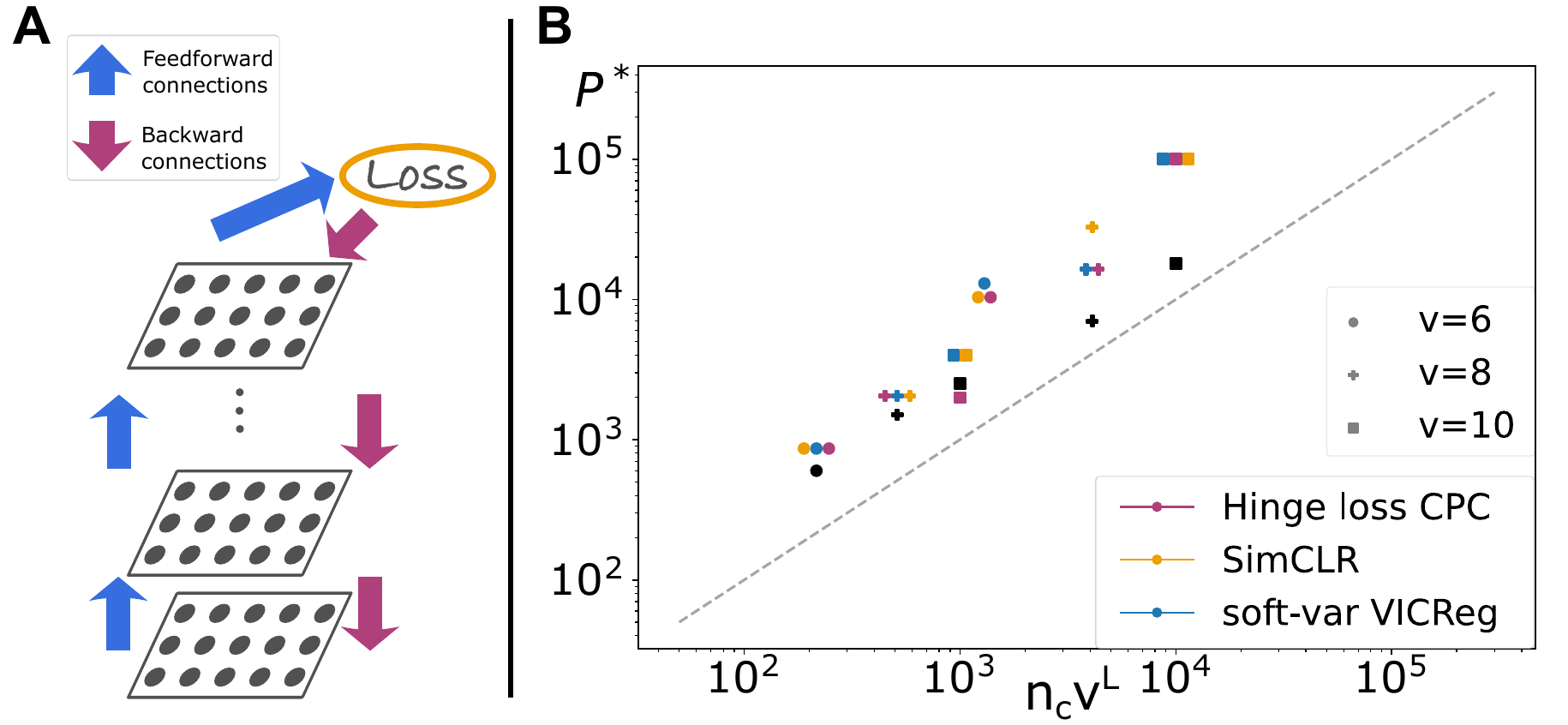}
    \caption{
	      {\bf Self-supervised learning algorithms solve  RHM tasks with end-to-end training.}
          {\bf A}. Schematic of forward and backward flow with the BP algorithm. The self-supervised loss function is applied at the output layer.
          {\bf B}.
          The minimum dataset size $P^*$ required to learn and generalize to the whole dataset (linear decoding error $\leq 10\%$ of random error), as a function of $D^*=n_c v^L$, for the RHMs with parameters $L=2, 3$ and $n_c = m = v=6, 8, 10$ (shape of symbols; color presents type of loss). CNNs trained with either Hinge-loss CPC (pink), soft-var VICReg (blue) or SimCLR (yellow) successfully form representations of the dataset with high data efficiency. Similar to local self-supervised learning, $P^*$ scales with the dimension of the hierarchical structure $D^*$. 
          \label{fig-ete-ssl} }
\end{figure}

The three self-supervised algorithms reviewed in Section \ref{sec-threeAlgos} -- two contrastive and one non-contrastive -- are tested on the RHM in an end-to-end fashion: the loss is computed at the final layer and BP is used to adapt the weights of the network (\autoref{fig-ete-ssl}A). We found that all three end-to-end self-supervised learning algorithms solve the RHM tasks with a data-efficiency similar to local self-supervised algorithms (\autoref{fig-ete-ssl}).

\section{Supplementary information: CLAPP without negative self-supervision \label{clapp-nc}} 
In the main results of the paper, CLAPP/Hinge-loss CPC uses both "positive" and "negative" self-supervision: when the loss is computed for a pair, the sign of $y$, which determines the sign of the plasticity and therefore whether prediction or contrast is applied, is positive if and only if the two encodings compared are from the same object.
While it is biologically conceivable that the notion of object continuity and permanence may be extended to object separation (in vision, that could be self-awareness of having made either a big saccade and thus having subsequent visual input from two different objects, or a fixation / micro-saccade and thus having subsequent visual input from a single object), 
this gives CLAPP a theoretical advantage over SimCLR, which uses random samples as negative pairs in the normalization, thus including pairs of encodings from the same object.
In SimCLR, optimizing the objective nonetheless finds separable representations for different objects because due to the expression of the softmax, the drive to increase the similarity of the positive pair is stronger than that to reduce the similarity of any single negative pair.
Especially for big values of $n_c$ the number of "negative pairs" that are actually same-object pairs becomes small compared to the number of real negative pairs, and the effect of false-negative pairs in the denominator can be neglected.

\begin{figure}
    \centering
   \includegraphics[width=10cm]{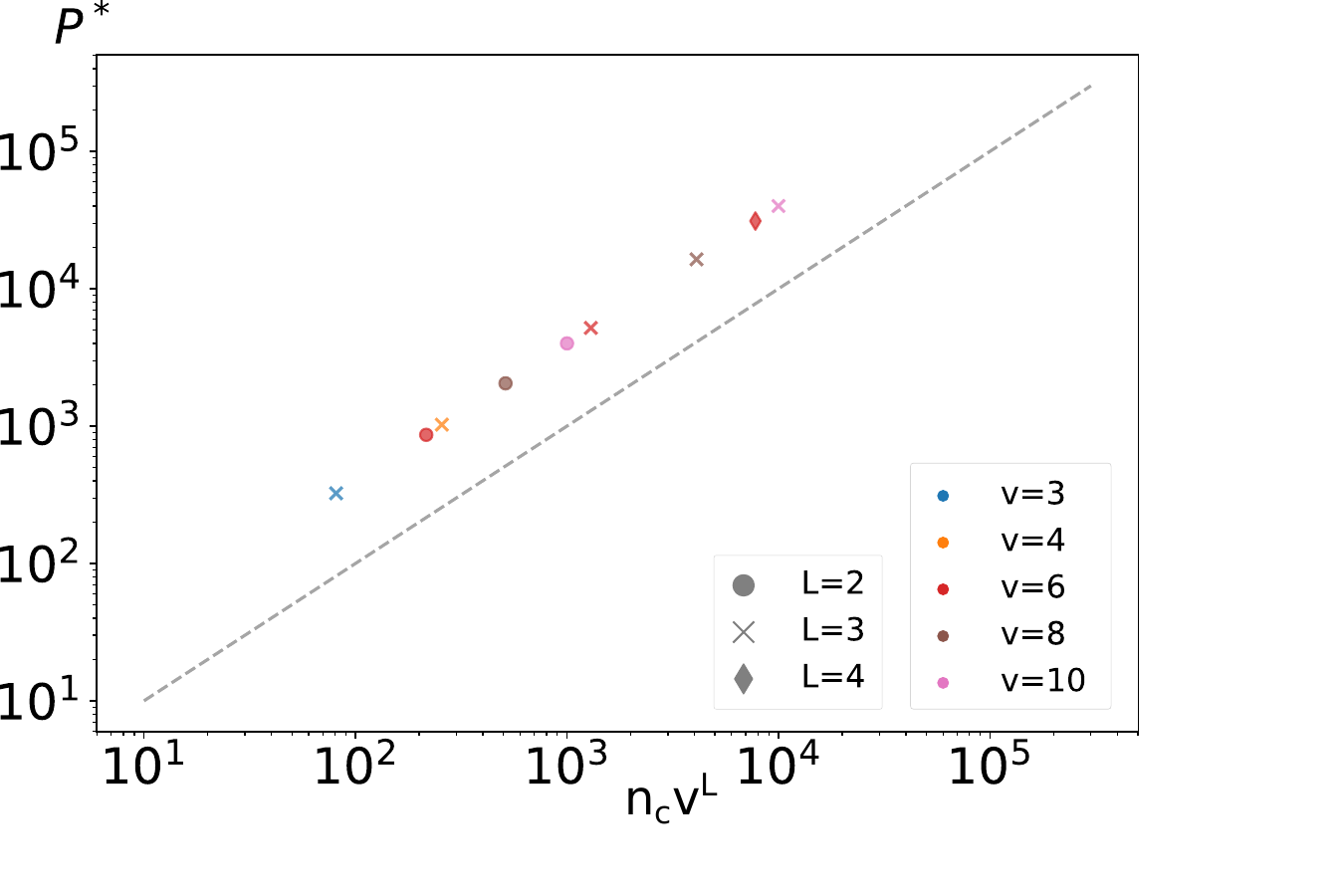}
    \caption{
	      {\bf  CLAPP solves the RHM in high data efficiency also in the absence of negative self-supervision.}
           As in \autoref{fig-lw}, the sample complexity scales linearly with $D^*$. Our data generated with a variant of CLAPP without negative self-supervision is plotted on the same scale as our Figures \ref{fig-ete-ssl} and \ref{fig-lw}, in the same format as the original Figure 3 of \citep{Cagnetta24}.
          \label{fig-clapp-nc} }
\end{figure}

In \autoref{fig-clapp-nc} we show that this reasoning also applies to CLAPP with no negative self-supervision: if "negative pairs" (for $y=-1$), are randomly drawn from \emph{all} pairs instead of from cross-object pairs only, 
the large number of true-negative pairs minimizes for positive pairs the "push-apart" effect compared to the "pull-together" effect, and the numerator is sufficient to counter-balance the "push-apart" in the denominator. Effectively, for positive pairs the "pull-together" effect dominates the representation of objects in this version of CLAPP (\autoref{fig-clapp-nc}). Indeed, this version of CLAPP is as powerful / data-efficient as standard CLAPP and other self-supervised algorithms.


\section{Supplementary information: Invariance of representations to synonymic exchange \label{clapp-inv}}

\begin{figure}
    \centering
   \includegraphics[width=\textwidth]{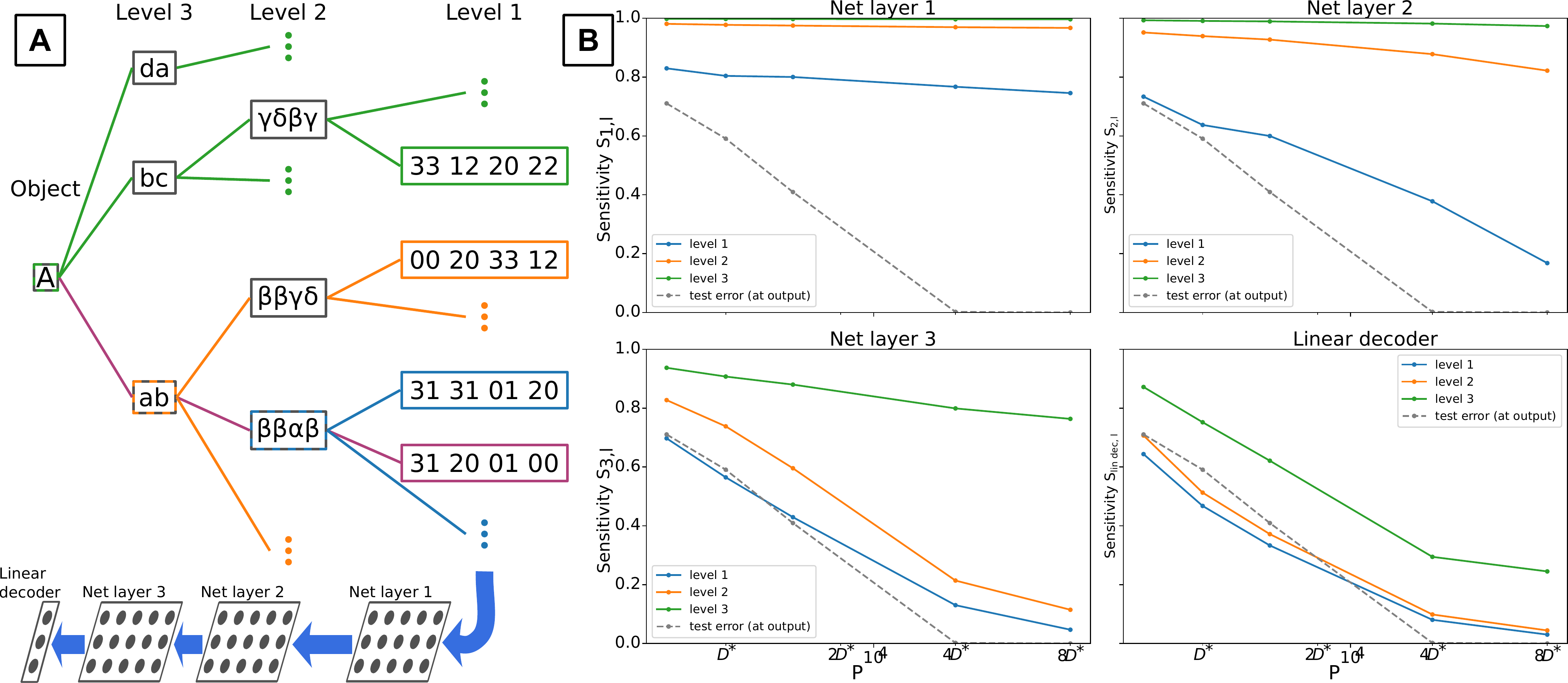}
    \caption{
	      {\bf Synonymic sensitivity of CLAPP representations.} 
          {\bf A} Example of RHM for  $L=3$, $m=3$, $v=4$. For the reference codeword in pink, we show examples of codewords with synonyms of level $l$ ($l=1$, blue; $l=2$, orange; $l=3$, green) exchanged. Any encodings of another object B are \emph{not} synonyms of the reference codeword for any level, because they do not share an encoding at any level.
          {\bf B}  Synonymic sensitivity (as defined in \autoref{eq-syn-sens}) of CLAPP representations as a function of size of data set. 
          Low values of sensitivity mean that the representations are invariant to the exchange of synonyms at the level indicated by the color
          at different layers.
          Data comes from an RHM with $L=3$, $m=n_c=v=8$. Each  panel shows sensitivity in on layer of the CLAPP-trained network, plus the usual linear decoder trained to readout object identity. As with supervised training \citep{Cagnetta24}, a CLAPP-trained model reaches synonymic invariance for level $l$ at layer $l+1$, and when the training set size reaches the critical size of a few $D^*$. Note that our model has as many layers as there are levels in the hierarchy; it is the linear decoder (equivalent of layer $L+1=4$) that achieves invariance to top-level synonymic exchange. Also, at suitable levels, representation invariance is achieved progressively as $P$ is increased, qualitatively parallel to the classification accuracy (grey dashed line).
           \label{fig-ss} }
\end{figure}

As in \citep{Cagnetta24} and in spite of the absence of top-down coordination of learning, we observe that CLAPP-trained models build representations that are invariant to the exchange of synonyms of level $l$ of the hierarchy, at layers $l+1$ and above, when the training set size reaches a few $D^*$.
This is illustrated in \autoref{fig-ss}.

\autoref{fig-ss}A shows an example RHM structure for  $L=3$ (and $m=3$, $v=4$). For a given codeword of reference (pink), we show in colors codewords in which synonyms of level $l$ have been exchanged, \emph{i.e.} that share their level $l+1$ encoding (in box with dashed color) with the reference codeword: $l=1$ blue, $l=2$ orange, $l=3$ green.

Very similar to \citep{Cagnetta24}, we define sensitivity $S_{k,l}$ of representations $\mathbf{z}^{(k)}$ at layer $k$ (of the NN) to the exchange of level-$l$ synonyms as:
\begin{equation} \label{eq-syn-sens}
    S_{k,l} = \frac{\left\langle \left\lVert\mathbf{z}^{(k)}(x) - \mathbf{z}^{(k)}(x^\prime) \right\rVert ^ 2 \right\rangle_{x, x^\prime \in \text{dataset} \text{ where } x \text{ and } x^\prime \text{ share the same encoding at level } l+1}}
    {\left\langle \left\lVert\mathbf{z}^{(k)}(x) - \mathbf{z}^{(k)}(x^\prime) \right\rVert ^ 2 \right\rangle_{x, x^\prime \in \text{dataset}}}
\end{equation}
For example, with the conventions of \autoref{fig-ss}A:
\begin{equation*}
    S_{k,l=2} = \frac{\left\langle\left\langle \left\lVert\mathbf{z}^{(k)}(\textcolor{mypink}{x}) - \mathbf{z}^{(k)}(\textcolor{myorange}{x^\prime}) \right\rVert ^ 2 \right\rangle_{\textcolor{myorange}{x^\prime}\text{ orange}} \right\rangle_{\textcolor{mypink}{x}\text{ pink}}}
    {\left\langle \left\lVert\mathbf{z}^{(k)}(x) - \mathbf{z}^{(k)}(x^\prime) \right\rVert ^ 2 \right\rangle_{x, x^\prime \in \text{dataset}}}
\end{equation*}

In \autoref{fig-ss}B, we show the sensitivity $S_{k,l}$ of CLAPP representations of layer $k$ to the exchange of level-$l$ synonyms for an instance of RHM with $L=3$, $m=n_c=v=8$: similar to the supervised case \citep{Cagnetta24}, CLAPP representations become invariant to the exchange of level-$l$ synonyms at layer $k=l+1$ and higher.

\printcredits

\bibliographystyle{cas-model2-names}

\newpage
\bibliography{bibliography2}

@Article{Aceituno24,
  author = 	 {P.V. Aceituno and S. \protect{de Haan} and R. Loidl and B.F. Grewe},
  title = 	 {Target Learning rather than Backpropagation Explains
Learning in the Mammalian Neocortex},
  journal = 	 {bioRxiv},
  year = 	 2024,
  volume = 	 {preprint},
  pages = 	 {DOI 10.1101/2024.04.10.588837}}

@Article{Akrout19,
  title={Deep learning without weight transport},
  author={Akrout, Mohamed and Wilson, Collin and Humphreys, Peter and Lillicrap, Timothy and Tweed, Douglas B},
  journal={Advances in neural information processing systems},
  volume={32},
  year={2019}
}

@Article{Bardes22,
  author = 	 {A. Bardes and J. Ponce and Y. \protect{LeCun}},
  title = 	 {Vicreg: variance-invariance-covariance re-
gularization for self-supervised learning},
  journal = 	 {ICLR},
  year = 	 2022,
  pages = 	 {arXiv:2105.04906v3}}

@ARTICLE{Bi01,
  author = {{G.-q.} Bi and {M.-m.} Poo},
  title = {Synaptic modification of correlated activity: Hebb's postulate revisited},
  journal = {Ann. Rev. Neurosci.},
  year = {2001},
  volume = {24},
  pages = {139-166}
}

@ARTICLE{Bliss03,
  author = {T.V.P. Bliss and G.L. Collingridge and R.G.M. Morris},
  title = {Long-term potentiation: enhancing neuroscience for 30 years - Introduction},
  journal = {Phil. Trans. R. Soc. Lond B: Biological Sciences},
  year = {2003},
  volume = {358},
  pages = {607-611}
}

@Article{Cagnetta24,
  author =      {F. Cagnetta and L. Petrini and U.M.  Tomasini and A. Favero and M. Wyart},
  title =      {How Deep Neural Networks Learn Compositional Data: The Random Hierarchy Model},
  journal =      {Phys. Rev. X},
  year =      2024,
  volume =      14,
  pages =      031001}

@inproceedings{Cagnetta2024b,
title={Towards a theory of how the structure of language is acquired by deep neural networks},
author={Francesco Cagnetta and Matthieu Wyart},
booktitle={The Thirty-eighth Annual Conference on Neural Information Processing Systems},
year={2024},
url={https://openreview.net/forum?id=NaCXcUKihH}
}

@inproceedings{Caron21,
  title={Emerging properties in self-supervised vision transformers},
  author={Caron, Mathilde and Touvron, Hugo and Misra, Ishan and J{\'e}gou, Herv{\'e} and Mairal, Julien and Bojanowski, Piotr and Joulin, Armand},
  booktitle={Proceedings of the IEEE/CVF international conference on computer vision},
  pages={9650--9660},
  year={2021}
}

@Article{Chen20,
  author = 	 {T. Chen and S. Kornblith and M. Norouzi and G. Hinton},
  title = 	 {A Simple Framework for Contrastive Learning
of Visual Representations.},
  journal = 	 {Proc. of the 37th Int. Conf. Mach. Learn. PMLR},
  year = 	 2020,
  volume = 	 119
}

@Article{Chen25,
  title={Self-Contrastive Forward-Forward Algorithm},
  author={Chen, Xing and Liu, Dongshu and Laydevant, J{\'e}r{\'e}mie and Grollier, Julie},
  journal={Nature Communications},
  volume={16},
  number={1},
  pages={5978},
  year={2025},
  publisher={Nature Publishing Group UK London}
}

@article{chizat19,
  title={On lazy training in differentiable programming},
  author={Chizat, Lenaic and Oyallon, Edouard and Bach, Francis},
  journal={Advances in neural information processing systems},
  volume={32},
  year={2019}
}

@Article{Clopath10,
  author = 	 {C. Clopath and  L. Busing and  E. Vasilaki and W. Gerstner},
  Title = 	 {Connectivity reflects coding: A model of voltage-based spike-timing-dependent-plasticity with homeostasis.},
  journal = 	 {Nature Neuroscience},
  year = 	 2010,
  volume = 	 {13},
  pages = 	 {344-352}}

@Article{Crick89,
  author = 	 {F.R. Crick},
  title = 	 {The recent excitement about neural networks},
  journal = 	 {Nature},
  year = 	 1989,
  volume = 	 337,
  pages = 	 {129-132}}

@Article{DiCarlo12,
  author = 	 {J.J. DiCarlo and D. Zoccolan and N.C. Rust},
  title = 	 {How Does the Brain Solve
Visual Object Recognition?},
  journal = 	 {Neuron},
  year = 	 2012,
  volume = 	 73,
  pages = 	 {415-434}}

@article{frenkel19,
  title={Learning without feedback: Direct random target projection as a feedback-alignment algorithm with layerwise feedforward training},
  author={Frenkel, Charlotte and Lefebvre, Martin and Bol, David},
  journal={stat},
  volume={1050},
  pages={3},
  year={2019}
}

@Article{Garrido23,
  author = 	 {Q. Garrido and Y. Chen and A. Bardes and L. Najman and Y. Lecun},
  title = 	 {On the duality between contrastive and non-contrastive self-supervised learning},
  journal = 	 {Elenvth Intern. Conf. Learning Repr.},
  year = 	 2023,
  note = 	 {DOI 10.48550/arXiv.2206.02574}}

@BOOK{Gerstner02,
  title = {Spiking Neuron Models: single neurons, populations, plasticity},
  publisher = {Cambridge University Press},
  year = {2002},
  author = {W. Gerstner and W. K. Kistler},
  address = {Cambridge UK}
}

@Article{Gerstner18,
  author = 	 {W. Gerstner and M. Lehmann and V. Liakoni and D. Corneil and J. Brea},
  title = 	 {Eligibility Traces and Plasticity on Behavioral Time Scales: Experimental Support of NeoHebbian Three-Factor Learning Rules},
  journal = 	 {Front. Neural Circ.},
  year = 	 2018,
  volume = 	 12,
  pages = 	 53,
  doi =	 {10.3389/fncir.2018.00053}}

@Article{Goltz21,
  author = 	 {J. Goltz and L. Kriener and A. Baumbach and S. Billaudelle and O. Breitweiser and B. Cramer and D. Dold and A.F. Kungl and W. Senn and J. Schemmel and K. Meier and M.A. Petrovici},
  title = 	 {Fast and energy-efficient neuromorphic deep learning with first-spike times},
  journal = 	 {Nature Machine Intelligence},
  year = 	 2021,
  volume = 	 3,
  pages = 	 {pages 823–835}
}

@Book{Goodfellow16,
  author = 	 {I. Goodfellow and Y. Bengio and A. Courville and Y. Bengio},
  title = 	 {Deep Learning},
  publisher = 	 {MIT Press, Cambridge Mass.},
  year = 	 2016}

@ARTICLE{Grossberg76,
  author = {S. Grossberg},
  title = {Adaptive pattern classification and universal recoding I: Parallel
	development and coding of neuronal feature detectors},
  journal = {Biol. Cybern.},
  year = {1976},
  volume = {23},
  pages = {121-134},
}

@Article{Halvagal23,
  author = 	 {M.S. Halvagal and F. Zenke},
  title = 	 {The combination of Hebbian and predictive plasticity learns invariant object representations in deep sensory networks},
  journal = 	 {Nat. Neurosci.},
  year = 	 2023,
  volume = 	 26,
  pages = 	 {1906-1915}}

@BOOK{Hebb49,
  title = {The {O}rganization of {B}ehavior},
  publisher = {Wiley},
  year = {1949},
  author = {D. O. Hebb},
  address = {New York}
}

@BOOK{Hertz91,
  title = {Introduction to the {T}heory of {N}eural {C}omputation},
  publisher = {Addison-Wesley},
  year = {1991},
  author = {J Hertz and A Krogh and R G Palmer},
  address = {Redwood City CA}
}

@misc{Hinton22forwardforward,
      title={The Forward-Forward Algorithm: Some Preliminary Investigations}, 
      author={Geoffrey Hinton},
      year={2022},
      eprint={2212.13345},
      archivePrefix={arXiv},
      primaryClass={cs.LG},
      url={https://arxiv.org/abs/2212.13345}, 
}

@ARTICLE{Hubel63,
  author = {Hubel, D.H. and Wiesel, T.N.},
  title = {RECEPTIVE FIELDS OF CELLS IN STRIATE CORTEX OF VERY YOUNG, VISUALLY
	INEXPERIENCED KITTENS},
  journal = {Journal of Neurophysiology},
  year = {1963},
  volume = {26},
  pages = {994--1002},
  number = {6},
  publisher = {Am Physiological Soc}
}

@ARTICLE{Hyvarinen00,
  author = {A. Hyv\protect{\"a}rinen and E. Oja},
  title = {Independent Component Analysis: algorithms and applications},
  journal = {Neural Networks},
  year = {2000},
  volume = {13},
  pages = {411--430},
  number = {4--5}
}

@Article{Hyvarinen98,
  author = 	 {A. Hyvarinen and E. Oja},
  title = 	 {Independent component analysis by general nonlinear Hebbian-like learning
rules.},
  journal = 	 {Signal Processing},
  year = 	 1998,
  volume = 	 64,
  pages = 	 {301-313}}

@Article{Illing19,
  author = 	 {B. Illing and W. Gerstner and J. Brea},
  title = 	 {Biologically plausible deep learning, but how far can we go with shallow networks?},
  journal = 	 {Neural Networks},
  year = 	 2019,
  volume = 	 {118},
  pages = 	 {90-101}
}

@inproceedings{Illing21,
 author = {B. Illing and J. Ventura and G. Bellec and W. Gerstner},
 booktitle = {Advances in Neural Information Processing Systems},
 editor = {M. Ranzato and A. Beygelzimer and Y. Dauphin and P.S. Liang and J. Wortman Vaughan},
 pages = {30365--30379},
 publisher = {Curran Associates, Inc.},
 title = {Local plasticity rules can learn deep representations using self-supervised contrastive predictions},
 url = {https://proceedings.neurips.cc/paper_files/paper/2021/file/feade1d2047977cd0cefdafc40175a99-Paper.pdf},
 volume = {34},
 year = {2021}
}

@article{Jacot18,
  title={Neural tangent kernel: Convergence and generalization in neural networks},
  author={Jacot, Arthur and Gabriel, Franck and Hongler, Cl{\'e}ment},
  journal={Advances in neural information processing systems},
  volume={31},
  year={2018}
}

@inproceedings{Laborieux24,
title={Improving equilibrium propagation without weight symmetry through Jacobian homeostasis},
author={Axel Laborieux and Friedemann Zenke},
booktitle={The Twelfth International Conference on Learning Representations},
year={2024},
url={https://openreview.net/forum?id=kUveo5k1GF}
}

@article{launay20,
  title={Direct feedback alignment scales to modern deep learning tasks and architectures},
  author={Launay, Julien and Poli, Iacopo and Boniface, Fran{\c{c}}ois and Krzakala, Florent},
  journal={Advances in neural information processing systems},
  volume={33},
  pages={9346--9360},
  year={2020}
}

@article{Lee19,
  title={Wide neural networks of any depth evolve as linear models under gradient descent},
  author={Lee, Jaehoon and Xiao, Lechao and Schoenholz, Samuel and Bahri, Yasaman and Novak, Roman and Sohl-Dickstein, Jascha and Pennington, Jeffrey},
  journal={Advances in neural information processing systems},
  volume={32},
  year={2019}
}

@Article{Lillicrap16,
  author = 	 {T.P. Lillicrap and D. Cownden and D.B. Tweed and C.J.Akerman},
  title = 	 {Random synaptic feedback weights support error backpropagation for deep learning},
  journal = 	 {Nature Communications},
  year = 	 2016,
  pages = 	 13276,
 anote =	 {doi:10.1038/ncomms13276}
 }

@Article{Lillicrap20,
  author = 	 {T.P. Lillicrap and A. Santoro and L. Marris and C.J. Akerman and G. Hinton},
  title = 	 {Backpropagation and the brain},
  journal = 	 {Nat. Rev. Neurosci.},
  year = 	 2020,
  volume = 	 21,
  pages = 	 {335-346}}

@Article{Maryada25,
  author = 	 {Maryada and C. DeLuca and A. Rubino and C. Wen and M. Cartiglia and I.I. Fodorut and M. Payvand and G. Indiveri},
  title = 	 {A canonical cortical electronic circuit for neuromorphic intelligence},
  journal = 	 {bioRxiv},
  year = 	 2025,
  note = 	 {DOI 2025.03.28.646019}}

@ARTICLE{Malenka04,
  author = {Robert C. Malenka and Mark F. Bear},
  title = {{LTP} and {LTD}: An Embarrassment of Riches},
  journal = {Neuron},
  year = {2004},
  volume = {44},
  pages = {5--21},
  anote = {the BEST recent REVIEW on Plasticity, because it is very readable and not too much molecular. CITE THIS ONE}
}

@Article{Max24,
  author = 	 {K. Max and L. Kriener and G.P. Garcia and T. Nowotny and I. Jaras and W. Senn and M.A. Petrovici},
  title = 	 {Learning efficient backprojections across cortical hierarchies in real time},
  journal = 	 {Nat. Mach. Intell.},
  year = 	 2024,
  volume = 	 6,
  pages = 	 {619-630}
}

@Article{Meulemans21,
  author = 	 {A. Meulemans and M.T. Farinha and J.G. Ordonez and P.V. Aceituno and J. Sacramento and B.J. Grewe},
  title = 	 {Credit Assignment in Neural Networks through
Deep Feedback Control},
  journal = 	 {ArXiv preprints},
  year = 	 2021,
  volume = 	 {ArXiv},
  pages = 	 {2106.07887}}

@ARTICLE{Meulemans22,
  title={Minimizing control for credit assignment with strong feedback},
  author={Meulemans, Alexander and Farinha, Matilde Tristany and Cervera, Maria R and Sacramento, Jo{\~a}o and Grewe, Benjamin F},
  journal={International Conference on Machine Learning},
  pages={15458--15483},
  year={2022},
  organization={PMLR}
}

@Article{Momeni25,
  author = 	 {A. Momeni and B. Rahmani and M. Mallejac and Philipp \protect{del Hougne} and Romain Fleury},
  title = 	 {Backpropagation-free training of deep physical neural networks},
  journal = 	 {Science},
  year = 	 2025,
  volume = 	 383,
  pages = 	 {1297–1303},
  note = 	 {DOI 10.1126/science.adi8474}}

@inproceedings{Nokland16,
 author = {N{\o}kland, Arild},
 booktitle = {Advances in Neural Information Processing Systems},
 editor = {D. Lee and M. Sugiyama and U. Luxburg and I. Guyon and R. Garnett},
 pages = {},
 publisher = {Curran Associates, Inc.},
 title = {Direct Feedback Alignment Provides Learning in Deep Neural Networks},
 url = {https://proceedings.neurips.cc/paper_files/paper/2016/file/d490d7b4576290fa60eb31b5fc917ad1-Paper.pdf},
 volume = {29},
 year = {2016}
}

@ARTICLE{Oja82,
  author = {E. Oja},
  title = {A simplified neuron model as a principal component analyzer},
  journal = {J. Mathematical Biology},
  year = {1982},
  volume = {15},
  pages = {267-273}
}

@Article{Oord19,
  author = 	 {A. van~den~Oord and Y. Li and O. Vinyals},
  title = 	 {Representation Learning with
Contrastive Predictive Coding},
  journal = 	 {arXiv},
  year = 	 2019,
  volume = 	 {arXiv},
  pages = 	 {1807.03748}
}

@Article{Patel15,
  author = 	 {A.B. Patel and T. Nguyen and R.G. Baraniuk},
  title = 	 {A Probabilistic Theory of Deep Learning},
  journal = 	 {arXiv preprints},
  year = 	 2015,
  volume = 	 {arXiv},
  pages =    {1504.00641}}

@Article{Pawlak10,
  author = 	 {V. Pawlak and J.R. Wickens and A. Kirkwood and J.N.D. Kerr},
  title = 	 {Timing is not everything: neuromodulation opens the {STDP} gate},
  journal = 	 {Front. Synaptic Neurosci.},
  year = 	 2010,
  volume = 	 2,
  pages = 	 146}

@inproceedings{Refinetti21,
  title={Align, then memorise: the dynamics of learning with feedback alignment},
  author={Refinetti, Maria and d’Ascoli, St{\'e}phane and Ohana, Ruben and Goldt, Sebastian},
  booktitle={International Conference on Machine Learning},
  pages={8925--8935},
  year={2021},
  organization={PMLR}
}

@incollection{Sacramento18,
title = {Dendritic cortical microcircuits approximate the backpropagation algorithm},
author = {J. Sacramento and R. {Ponte Costa} and Y. Bengio and W. Senn},
booktitle = {Advances in Neural Information Processing Systems 31},
editor = {S. Bengio and H. Wallach and H. Larochelle and K. Grauman and N. Cesa-Bianchi and R. Garnett},
pages = {8721--8732},
year = {2018},
publisher = {Curran Associates, Inc.},
url = {http://papers.nips.cc/paper/8089-dendritic-cortical-microcircuits-approximate-the-backpropagation-algorithm.pdf}
}

@inproceedings{Salvatori24,
  title = "A Stable, Fast, and Fully Automatic Learning Algorithm for Predictive Coding Networks",
  author = "Salvatori, Tommaso and Song, Yuhang and Yordanov, Yordan and Millidge, Beren and Xu, Zhenghua and Sha, Lei and Emde, Cornelius and Bogacz, Rafal and Lukasiewicz, Thomas",
  year = "2024",
  booktitle = "Proceedings of the 12th International Conference on Learning Representations, ICLR 2024, Vienna, Austria, 7--11 May 2024",
}

@Article{Scellier17,
  author = 	 {B. Scellier and Y. Bengio},
  title = 	 {Equilibrium propagation: Bridging the gap between energy-based models and backpropagation},
  journal = 	 {Front. Comput. Neurosci.},
  year = 	 2017,
  volume = 	 11,
  pages = 	 24}

@ARTICLE{Sjostrom01,
  author = {P.J. Sj{\"o}str{\"o}m and G.G. Turrigiano and S.B. Nelson},
  title = {Rate, timing, and cooperativity jointly determine cortical synaptic
	plasticity},
  journal = {Neuron},
  year = {2001},
  volume = {32},
  pages = {1149-1164}
}

@ARTICLE{Sprekeler07,
  author = {Henning Sprekeler and Christian Michaelis and Laurenz Wiskott},
  title = {Slowness: An objective for spike-timing-plasticity?},
  journal = {PLoS Computational Biology},
  year = {2007},
  volume = {3},
  pages = {e112},
  number = {6}
}

@inproceedings{Srinivasan24,
  title={Forward Learning with Top-Down Feedback: Empirical and Analytical Characterization},
  author={Srinivasan, Ravi Francesco and Mignacco, Francesca and Sorbaro, Martino and Refinetti, Maria and Cooper, Avi and Kreiman, Gabriel and Dellaferrera, Giorgia},
  booktitle={ICLR},
  year={2024}
}

@Article{Urbanczik14,
  author = 	 {R. Urbanczik and W. Senn},
  title = 	 {Learning by the dendritic prediction of somatic spiking},
  journal = 	 {Neuron},
  year = 	 2014,
  volume = 	 81,
  pages = 	 {521-528}}

@ARTICLE{Whittington17,
  author = {J.C.r. Whittington and R. Bogacz},
  title = {An approximation of the error
backpropagation algorithm in a predictive coding network with local hebbian
synaptic plasticity},
  journal = {Neural Comput.},
  year = {2017},
  volume = {29},
  pages = {1229-1262}
}

@Article{Williams19,
  author = 	 {L.E. Williams and A. Holtmaat},
  title = 	 {Higher-Order Thalamocortical Inputs Gate Synaptic Long-
Term Potentiation via Disinhibition.},
  journal = 	 {Neuron},
  year = 	 2019,
  volume = 	 101,
  pages = 	 {91-102}
}

@Article{Yamins16,
  author = 	 {D.L.K. Yamins and J.J. \protect{DiCarlo}},
  title = 	 {Using goal-driven deep learning models to understand sensory cortex},
  journal = 	 {Nat. Neurosci.},
  year = 	 2016,
  volume = 	 19,
  pages = 	 {356-365}}

@misc{Mossel18,
	title = {Deep {Learning} and {Hierarchal} {Generative} {Models}},
	url = {http://arxiv.org/abs/1612.09057},
	doi = {10.48550/arXiv.1612.09057},
	urldate = {2025-05-15},
	publisher = {arXiv},
	author = {Mossel, Elchanan},
	month = sep,
	year = {2018},
	note = {arXiv:1612.09057 [cs]},
}

@article{Mhaskar17,
	title = {When and {Why} {Are} {Deep} {Networks} {Better} {Than} {Shallow} {Ones}?},
	volume = {31},
	copyright = {Copyright (c)},
	issn = {2374-3468},
	url = {https://ojs.aaai.org/index.php/AAAI/article/view/10913},
	doi = {10.1609/aaai.v31i1.10913},
	language = {en},
	number = {1},
	urldate = {2025-05-15},
	journal = {Proceedings of the AAAI Conference on Artificial Intelligence},
	author = {Mhaskar, Hrushikesh and Liao, Qianli and Poggio, Tomaso},
	month = feb,
	year = {2017},
	note = {Number: 1},
}

@inproceedings{Cagnetta23,
	title = {What {Can} {Be} {Learnt} {With} {Wide} {Convolutional} {Neural} {Networks}?},
	url = {https://proceedings.mlr.press/v202/cagnetta23a.html},
	language = {en},
	urldate = {2025-05-15},
	booktitle = {Proceedings of the 40th {International} {Conference} on {Machine} {Learning}},
	publisher = {PMLR},
	author = {Cagnetta, Francesco and Favero, Alessandro and Wyart, Matthieu},
	month = jul,
	year = {2023},
	note = {ISSN: 2640-3498},
	pages = {3347--3379},
}



\end{document}